\documentclass[english,11pt]{article}
\usepackage{geometry}
\pdfoutput=1
\usepackage{authblk}

\usepackage{graphicx}
\usepackage{hyperref}
\usepackage[table]{xcolor}
\usepackage{color}
\usepackage{dirtree}
\usepackage{subfig}
\usepackage[modulo, switch]{lineno}

\usepackage{amsmath}
\usepackage{graphicx}        
\usepackage{multicol}        
\usepackage{lscape}
\usepackage{float}
\usepackage{url}
\usepackage{caption,subfig}
\usepackage{graphicx}
\usepackage{amsmath}
\usepackage{amssymb}
\usepackage{color}
\usepackage{blkarray}

\usepackage{algorithm}
\usepackage{algpseudocode}

\usepackage{tikz,graphicx}
\usepackage{url}
\usepackage{fancyhdr}
\usepackage{mathtools}
\usepackage{relsize}
\usepackage{cite}

\usetikzlibrary{positioning,arrows.meta,calc,decorations.pathreplacing}
\usepackage{xifthen}

\graphicspath{{pictures/}} 

\def\ZENO{{\sc ZenoTravel}}

\def\MULTIZENO{{\sc MultiZenoTravel}}
\def\ZENOSOLVER{{\sc ZenoSolver}}

\renewcommand{\vec}[1]{\mathbf{#1}}
\newcommand{\weakdom}{\leq} 
\newcommand{\reals}[1]{\mathbb{R}}
\newcommand{\ints}[1]{\mathbb{N}}

\def\C++{%
    {\sffamily C\kern-.1667em\raise.50ex\hbox{\smaller[3]{++}}}%
\spacefactor1000 }

\makeatletter
\def\captionof#1#2{{\def\@captype{#1}#2}}
\makeatother

\definecolor{nojblue}{rgb} {0.0, 0.2, 0.6}
\definecolor{nojgreen}{rgb} {0.0, 0.4, 0.1}
\definecolor{nojmagenta}{rgb} {0.6, 0.1, 0.6}

\hypersetup{
    colorlinks=true,
    linktoc=all, 
    pdftitle={Quemy, Schoenauer, Dreo: MultiZenoTravel: a Tunable Benchmark for Multi-Objective Planning with Known Pareto Front},
    linkcolor=nojmagenta,  
    citecolor=nojgreen, 
    filecolor=nojblue, 
    urlcolor=nojblue, 
}

\title{{\sc MultiZenoTravel}: a Tunable Benchmark for Multi-Objective Planning with Known Pareto Front}

\author[1,2,*]{Alexandre Quemy}
\author[3,4]{Marc Schoenauer}
\author[5,6]{Johann Dreo}

\affil[1]{YData Lab Inc., Seattle, United States, \texttt{<alexandre.quemy@ydata.ai>} (${}^*$corresponding author)}
\affil[2]{Poznan University of Technology, Poznan, Poland}

\affil[3]{Inria Saclay, Orsay, France, \texttt{<marc.schoenauer@inria.fr>}}
\affil[4]{Universite Paris-Saclay, Orsay, France}

\affil[5]{Computational Systems Biomedicine laboratory, Department of Computational Biology, Institut Pasteur, Paris, France, \texttt{<johann.dreo@pasteur.fr>}}
\affil[6]{Bioinformatics and Biostatistics hub, Université Paris Cité, Institut Pasteur, Paris, France}

\date{\today}

\begin{document}
\maketitle

\begin{abstract}
Multi-objective AI planning suffers from a lack of benchmarks exhibiting known Pareto Fronts. 
In this work, we propose a tunable benchmark generator, together with a dedicated solver that provably computes the true Pareto front of the resulting instances.
First, we prove a proposition allowing us to characterize the optimal plans for a constrained version of the problem, and then show how to reduce the general problem to the constrained one. 
Second, we provide a constructive way to find all the Pareto-optimal plans and discuss the complexity of the algorithm. 
We provide an implementation that allows the solver to handle realistic instances in a reasonable time.
Finally, as a practical demonstration, we used this solver to find all Pareto-optimal plans between the two largest airports in the world, considering the routes between the 50 largest airports, spherical distances between airports and a made-up risk.

\end{abstract}



\setcounter{page}{0}
\maketitle

\section{Introduction}
\label{sec:intro}

The progress of algorithmics, the availability of more \& more data and the dramatic increase of computational power drive a fast-pace evolution of the artificial intelligence (AI) field.
As part of this change, the need to assess the performances of computational methods and to compare their merits is crucial.
Many of the core fields of AI have set up standard benchmarks and competitions, in order to complement expert knowledge and analysis.
In that regard, the automated planning community is at the edge of the state of the art, with the well-known International Planning Competition (IPC), hosting a large benchmark and using a common definition language.

A deterministic planning problem consists in selecting a sequence of actions ---a plan--- having an effect on a state, so that applying the plan on an initial state allows to reach a goal (partial) state, while optimizing a function of the plan's value.
This function is generally the total duration to reach the goal (the {\em makespan}), each action having a duration.
However, the value function may very well represent another aspect of the problem, such as the cost of the plan, the energy it requires or the uncertainty produced by the actions.

In realistic problems, it is very often the case that several such {\em objective functions} exists and are contradictory.
For example, a short plan may be costly, while a cheap plan may take a long time.
In such a setting, using a linear combination of those objective functions falls back to introducing a bias about the preferences of the operational user.
However, preferences cannot always be modelled easily in practice, e.g. the user may decide based on political information. In addition, the knowledge of the feasible compromises between the objectives may, in fact, influence the decision maker simply because it gives additional information about the problem itself. For instance, if the two objectives are a cost and a risk, the decision maker might revise its risk appetite knowing that the increase of the risk by 1\% beyond its initial acceptable risk threshold can lead to a 50\% cost decrease.

This calls for the use of multi-objective optimization, where the problem is actually modelled with several objective functions, and the output of the solver is a set of solutions that are non-dominated by other solutions, regarding the objectives.
The weak dominance of a $d$-dimensional point 
$\vec{a}$ over point $\vec{b}$ is defined as $\vec{a} \weakdom \vec{b} \iff \vec{a}^d \leq \vec{b}^d \ \forall d \in \ints{}^+$.
A set of point $P$ is then defined as Pareto-optimal if it dominates every other points in $X$, all its points being non-dominated by each others:
$\vec{p} \weakdom \vec{x} \land \nexists \vec{q} \mid \vec{q} \weakdom \vec{p} \;\forall \vec{p}, \vec{q} \in P,\; \vec{x} \in X$.
The output of solving such a problem is thus a set of solutions ordered as a ``Pareto front''.
That is, for a problem with two objective functions, a set of points ordered along a monotonic function.
Operationally, the user is still in charge of taking the final decision,
but the complexity of their decision has been drastically reduced to a $d-1$ dimensional problem.

While real-world problems are often multi-objective in nature,
few work actually consider their study in the automated planning area.
The well known IPC do not proposes benchmark for such 
problems~\cite{vallati_chrpa_mccluskey_2018,inp_Cenamor2019} to this date.
In fact, PDDL 3.0 explicitly offered hooks for several objectives~\cite{GEREVINI2009619},
but the only organized competition tracks concerned aggregated objectives,
tracks which were canceled in 2011.

Despite various work on benchmark 
generation~\cite{a_Lagriffoul2018,tr_Akguen2020,inp_Balyo2022}
and extension to other planning problems~\cite{tr_Pellier2021},
no truly multi-objective planning problem has been proposed apart from our line of work.

\subsection{Previous Works}

We have previously proposed a problem instance generator for such multi-objective planning problems~\cite{Schoenauer2006,khouadjia2013multi,quemy:hal-01109777},
extending the \ZENO{}~\cite{penberthy1994temporal} problem with an additional objective.
Such problem sets are crucial for benchmarking optimization algorithms,
especially when the optimum is known, as it allows for a rigorous comparison of the performances of solvers.

The original \ZENO{} problem~\cite{penberthy1994temporal} involves planes moving passengers between cities, while taking care of their fuel.
Actions such as flying, boarding, deplaning or refueling take various time to complete, and a plane cannot fly without fuel.
The objective is to minimize the makespan, while honoring passengers’ destinations.

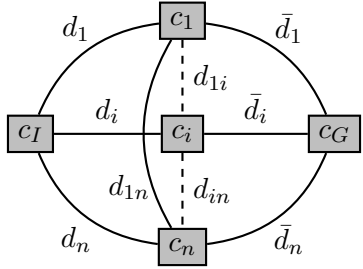
\begin{figure}[htbp!]
\centering
 \begin{tikzpicture}[thick,scale=1]
  \node[draw, fill=black!25] (0) at (-2,0) {$c_I$};
  \node[draw, fill=black!25] (1) at (0,1.5) {$c_1$};
  \node[draw, fill=black!25] (2) at (0,0) {$c_i$};
  \node[draw, fill=black!25] (3) at (0,-1.5) {$c_n$};
  \node[draw, fill=black!25] (4) at (2,0) {$c_G$};
 
  \draw[thick] (0) to[bend left] (1);
  \draw[thick] (0)--(2) node[midway, above]{$d_i$};
  \draw[thick, bend left] (0) to[bend right] (3);
  
  \draw[thick] (1) to[bend left] (4);
  \draw[thick] (2)--(4) node[midway, above]{$\bar d_i$};
  \draw[thick] (3) to[bend right] (4);
  
  \draw[dashed] (1)--(2) node[midway, right]{$d_{1i}$};
  \draw[dashed] (2)--(3) node[midway, right]{$d_{in}$};
  \draw[thick] (1) to[bend right] (3);
     
  \node[] (A) at (-1.4,1.4) {$d_1$};
  \node[] (A) at (1.4,1.4) {$\bar d_1$};
  \node[] (A) at (-1.4,-1.4) {$d_n$};
  \node[] (A) at (1.4,-1.4) {$\bar d_n$};
  \node[] (A) at (-0.7,-0.7) {$d_{1n}$};
  
\end{tikzpicture}
\caption{\label{Figureinstance}A schematic view of a non-symmetric clique \MULTIZENO\ problem. }
\end{figure}

The proof-of-concept of the \MULTIZENO{} problem is based on a simplified \ZENO{} model~\cite{Schoenauer2006}.
In this model, there are five connected cities (see Figure~\ref{Figureinstance}),
planes may transport only one passenger at a time and there is only one flying speed.
The main addition to the problem is that an additional objective is attached to all actions.
This second objective is either a {\em cost}, which is additive (each plane has to pay the corresponding tax every time it lands in that city)
either a {\em risk} (for which the maximal value encountered during the complete execution of a plan is to be minimized).
In this first instance, three passengers can be moved across the cities.

The second version of the \MULTIZENO{} problem~\cite{khouadjia2013multi} builds up on the proof-of-concept, allows for 3, 6 and 9 passengers,
and a 3-to-2 passengers-to-planes ratio, which makes up for very small instances,
due to the combinatorial explosion of the solution space.

In \cite{quemy:hal-01109777}, we introduce an algorithm to compute the true Pareto fronts
of very large instance in reasonable time,
and the first version of the \ZENOSOLVER{} software is described
(see Section~\ref{sec:zenosolver} for further details).
The article also provides a few typical instances that exhibit very different shapes of Pareto Fronts,
for different levels of complexity. Unfortunately, this works suffer from two unrealistic assumptions. First, the distances are assumed to be symmetric around the central cities, that is to say $f\forall i, d_i = \bar d_{i}$. Second, the proof of the proposition which allows us to find Pareto-optimal plans relies on the following unrealistic assumption: 
\begin{equation*}
    \forall (i, j) \in [1, n]^2, d_i + d_j < d_{ij}
\end{equation*}
In other words, none of the instances generated under this assumption would respect a triangular inequality. Even if a benchmark does not necessarily have to be realistic by nature, this particular assumption drastically restrains the extrapolation of the performances of a solver observed on the benchmark to real life problems.

\subsection{Our Contribution}

In this work, we introduce a constructive way to find the Pareto-optimal solutions for the \MULTIZENO{} problem. Our first contribution is to generalize from the original symmetric clique problem introduced in~\cite{Schoenauer2006} to a non-symmetric clique version and to a version with no particular assumption on the graph. 

As a second contribution, we provide a way to characterize the Pareto-optimal plans for all versions of the problem, leading to a constructive algorithm to find the Pareto-optimal solution for any instance. In particular, we would like to insist on the fact that this paper is not an extension of \cite{quemy:hal-01109777} to a more generic case. In fact, in \cite{quemy:hal-01109777} the unrealistic aforementioned assumption contradicts the triangular inequality assumption made in this paper. In addition, even the technical implementation of the algorithm have been completely revised. The only intersection between the papers is contained in Section \ref{sec:ppp} and concerns how we define and count the potential and admissible Pareto Optimal Plans.

In addition, we present the \ZENOSOLVER, a C++ implementation of the algorithm to solve \MULTIZENO{}. It can output the instance definition in PDDL such that the generated instance can easily be used by other solvers. We demonstrate how to generate instances with different behavior by tuning the input parameters. We provide a theoretical and empirical study of the performances of \ZENOSOLVER. 

Although we think that the primary utility of \ZENOSOLVER~is to generate benchmarks with known Pareto Front to study other solvers' properties and behaviors, we provide a demonstration of the \ZENOSOLVER~on a problem using real data. This had been made possible because of the extension of the solver capabilities from the symmetric clique problem to any arbitrary graph.

\subsection{Outline}

The plan of this paper is as follows: in Section \ref{sec:multizeno}, we define three versions of the \MULTIZENO~problem.
Section \ref{sec:symmetric} is dedicated to solve the symmetric \MULTIZENO~problem, followed by Section \ref{sec:non_symmetric} that focuses on the non-symmetric version of the problem.
In Section \ref{sec:general_multizeno}, we show how any instance of the general \MULTIZENO~problem can be reduced to a non-symmetric version via a polynomial-time reduction.

In Section \ref{sec:ppp}, we detail how to construct Pareto optimal plans for the non-symmetric problem.
In Section \ref{sec:zenosolver}, we introduce the \ZENOSOLVER{}, a \C++ implementation of the algorithm described in the previous sections.

Finally, in Section \ref{sec:application}, we study an (almost) realistic application for \MULTIZENO{} and \ZENOSOLVER{} using the Openflight database to find the optimal routes between the two largest airports in the world.

\section{\MULTIZENO\ problems}
\label{sec:multizeno}

In this Section, we introduce three versions of the \MULTIZENO{} problem: the symmetric clique \MULTIZENO, the non-symmetric clique \MULTIZENO{} and the general \MULTIZENO.
First, we will prove a proposition that characterizing the non-optimal plans for the symmetric clique \MULTIZENO~and therefore, helps building optimal plans. Then, we show that the for the non-symmetric version, the proposition still holds. Finally, we show that we can reduce the general \MULTIZENO to a clique version, and thus, still build optimal plans.

\subsection{Instances}
Let us introduce some notations related to the planning problem briefly presented in the introduction: a non-symmetric clique \MULTIZENO\ instance (Figure \ref{Figureinstance}) is defined by the following elements:
\begin{itemize}
\item $n$ central cities, organized as a clique in which every node is connected to $C_I$ and $C_G$, respectively the initial city and the goal city.
\item $c\in ({\mathbb{R}^+})^n$, where $c_i$ is the cost for landing in $C_i$.
\item $D \in ({\mathbb{R}^+})^{n \times n}$, where $d_{ij}$ is the flying time between $C_i$ and $C_j$.
\item $d \in ({\mathbb{R}^+})^n$, where $d_i$ is the flying time between $C_I$ and $C_i$.
\item $\bar d \in ({\mathbb{R}^+})^n$, where $\bar d_i$ is the flying time between $C_i$ and $C_G$.
\item $p$ planes, initially in $C_I$, that have a capacity of a unique person.
\item $t$ persons, initially in $C_I$.
\end{itemize}
The goal of \MULTIZENO\ is to carry all $t$ persons, initially in $c_I$, to $c_G$ using $p$ planes, minimizing both the makespan and the cost of the plan.

\noindent
Without loss of generality, all pairs $(d_i, c_i)$ are assumed to be pairwise distinct.
Otherwise, the 2 cities can be ``merged'' and the resulting $n-1$ cities problem is equivalent to the original $n$ cities problem, as there exist no city capacity constraints.
Finally, we only consider cases where $t \geq p$, as the problem is otherwise trivial.

An instance of the symmetric clique \MULTIZENO~problem is an instance such that $d = \bar d$. A general \MULTIZENO~instance is an instance such that the $n$ central cities are organized as an arbitrary graph.

\begin{figure}[!h]
$$
\begin{matrix}
   p_1: & C_I & \overset{t_1} \to C_4 & \to C_G & \to C_2 & \overset{t_2} \to C_G & \\
   p_2: & C_I & \overset{t_2} \to C_2 & \to C_I & \overset{t_3} \to C_3 & \overset{t_3} \to C_4 & \to C_G & & 
\end{matrix}
$$
\caption{Example of an admissible plan to transport 3 travelers via 2 planes. This representation indicates the successive actions for each plane. We noted by $\overset{t_i} \to$ a flight with a traveler.}
\label{fig:plan_example}

\end{figure}

 Figure \ref{fig:plan_example} illustrates an admissible solution. Note that the makespan for a plan is not necessarily the largest sum of the flights' duration as some planes might have to wait for others. This could be the case for $p_1$ waiting for $t_2$ in $C_2$.

\subsection{Symmetric \MULTIZENO}
\label{sec:symmetric}

The method to find the Pareto Front of any general \MULTIZENO~instance consists of three steps. First, we provide an efficient algorithm to find the Pareto Front for any symmetric clique \MULTIZENO~instance. Then, we show that there exists one particular case in which the algorithm does not work for the non-symmetric version of the problem. However, we provide a way to easily transform the instance such that a slightly modified version of the algorithm can find the Pareto Front. Last, we provide an algorithm to transform any instance of the general \MULTIZENO~problem into a non-symmetric clique version.


The following proposition is the cornerstone of the method to identify and construct the Pareto Set of any instance:\\

\noindent
{\bf Proposition I}: Pareto-optimal plans are plans where exactly $2t-p$ (possibly identical) central cities are used by a plane.\\

In particular, Proposition I will be proven only for the symmetric case as we will show there is one corner-case to the demonstration for the non-symmetric version. We will overcome this corner-case by a slight modification of the solver's main algorithm.

\noindent
For the rest of this paper, we make the following triangular inequality assumption:\\

\noindent
{\bf Assumption}: $\forall (i, j, k) \in ([1,n]\cup \{I,G\})^3, ~ d_{ij} + d_{jk} \geq d_{ik}$ \hfill (A$\Delta$)\\

The goal of this section is to establish the proof of Proposition I for the Symmetric case. For that purpose, we will first determine a couple of properties, mostly deduced from (A$\Delta$), to restrict the movements of the planes in Pareto Optimal plans to four different patterns.\\

\noindent
{\bf Property I}:
\begin{enumerate}
\item A plane flies from $C_I$ with a passenger and to $C_I$ empty.
\item A plane flies from $C_G$ empty and to $C_G$ with a passenger.
\item A plane does not fly two times in a row between central cities with or without a passenger.
\end{enumerate}

\noindent
{\bf Proof}: Straightforward consequences of (A$\Delta$).\hfill $\square$\\

\noindent
{\bf Corollary I}: All the planes finish their respective sequence in $C_G$.\\

\noindent
{\bf Proof}: If a plane finishes in $C_I$, it cannot arrives full from a central city due to Property~I.1 nor empty because the last move would be useless. If the plane finishes in $C_i$ and came empty, it cannot comes from $C_I$ due to Property I.1. It cannot come from $C_j$ another central city or $C_G$ because the movement would be useless. If the plane finishes in $C_i$ and came full, another plane will have to carry the passenger to $C_G$ and thus, it would be at least as fast to go directly to $C_G$ with the inital plane. \hfill $\square$\\

From those observations, we deduce the only four possible patterns that a plane can perform, denoted by $A$, $\bar A$, $B$, $\bar B$. More precisely, if a plane does perform another pattern than these ones, the plan is dominated by the plan corrected in such a way that it respects Property~I because the second plan uses less cities (the makespan might be the same but no longer as insured by the triangular inequality). \\

\noindent
\begin{figure}
\centering
\begin{minipage}{.2\textwidth}
 \begin{tikzpicture}[thick,scale=0.4]
  \node[draw, fill=black!25] (0) at (-3.5,-2) {\large $c_I$};
  \node[draw, fill=black!25] (1) at (0,-2) {\large $c_{i_1}$};
  \node[draw, fill=black!25] (2) at (0,0) {\large $c_{i_2}$};
  \node[draw, fill=black!25] (3) at (0,2) {\large $c_{i_{2\theta}}$};
  \node[draw, fill=black!25] (4) at (3.5,2) {\large $c_G$};
 
  \draw[thick, ->] (0)--(1) node[midway, above]{.};
  \draw[thick, ->] (1)--(2) node[midway, right ]{};
  \draw[dotted, ->] (2)--(3) node[midway, above ]{};
  \draw[thick, ->] (3)--(4) node[midway, above ]{.};

 \end{tikzpicture}

\end{minipage}
\hspace{3cm}
\begin{minipage}{.2\textwidth}
 \begin{tikzpicture}[thick,scale=0.4]
  \node[draw, fill=black!25] (0) at (3.5,-2) {\large $c_G$};
  \node[draw, fill=black!25] (1) at (0,-2) {\large $c_{i_1}$};
  \node[draw, fill=black!25] (2) at (0,0) {\large $c_{i_2}$};
  \node[draw, fill=black!25] (3) at (0,2) {\large $c_{i_{2\theta}}$};
  \node[draw, fill=black!25] (4) at (-3.5,2) {\large $c_I$};
 
  \draw[thick, ->] (0)--(1) node[midway, above]{};
  \draw[thick, ->] (1)--(2) node[midway, right ]{.};
  \draw[dotted, ->] (2)--(3) node[midway, above ]{};
  \draw[thick, ->] (3)--(4) node[midway, above ]{};
 
 \end{tikzpicture}
\end{minipage}
\end{figure}

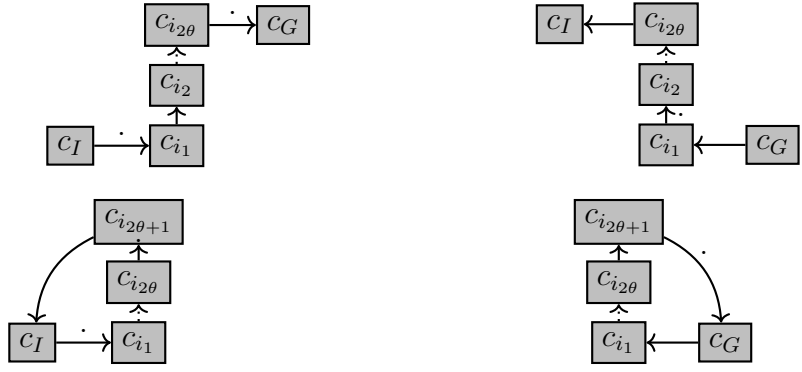
\begin{figure}
\centering
\begin{minipage}{.2\textwidth}
 \begin{tikzpicture}[thick,scale=0.4]
  \node[draw, fill=black!25] (0) at (-3.5,-2) {\large $c_I$};
  \node[draw, fill=black!25] (1) at (0,-2) {\large $c_{i_1}$};
  \node[draw, fill=black!25] (2) at (0,0) {\large $c_{i_{2\theta}}$};
  \node[draw, fill=black!25] (3) at (0,2) {\large $c_{i_{2\theta+1}}$};
 
  \draw[thick, ->] (0)--(1) node[midway, above]{.};
  \draw[dotted, ->] (1)--(2) node[midway, right ]{};
  \draw[thick, ->] (2)--(3) node[midway, above ]{.};
  \draw[thick, ->] (3) to[bend right] (0);
 
 \end{tikzpicture}
\end{minipage}
\hspace{4cm}
\begin{minipage}{.2\textwidth}
  \begin{tikzpicture}[thick,scale=0.4]
  \node[draw, fill=black!25] (0) at (3.5,-2) {\large $c_G$};
  \node[draw, fill=black!25] (1) at (0,-2) {\large $c_{i_1}$};
  \node[draw, fill=black!25] (2) at (0,0) {\large $c_{i_{2\theta}}$};
  \node[draw, fill=black!25] (3) at (0,2) {\large $c_{i_{2\theta+1}}$};
 
  \draw[thick, ->] (0)--(1) node[midway, above]{};
  \draw[dotted, ->] (1)--(2) node[midway, right ]{};
  \draw[thick, ->] (2)--(3) node[midway, above ]{};
  \draw[thick, ->] (3) to[bend left] (0);
  \node[] (A) at (2.8,1) {$.$};
 
 \end{tikzpicture}
 \end{minipage}
 \caption{\label{M3} On top, Pattern $A$ and $\bar A$. In the bottom, $B$ and $\bar B$. The dots above the arrows indicate a flight with a passenger.}
\end{figure}

\noindent
We denote by $|X|$ the number of the patterns $X \in \{A, \bar A, B, \bar B\}$ in a given plan and call {\it multiplicity} the number $\theta$ associated to a specific pattern execution.
Depending on the pattern, the number of cities involved is either even or odd, so as to respect Property I.
Using Property I and the triangular inequality, we deduce the following property on the cardinal of each patterns in potential Pareto plans:\\

\noindent
{\bf Property II}: If a plan does not respect the following constraints, it is dominated:
\begin{enumerate}
  \item $|A| + |B| = t$
  \item $|A| + |\bar B| = t$
  \item $|B| = |\bar B|$
  \item $|A| = |\bar A| + p$
\end{enumerate}
~\\\noindent
{\bf Proof}: The pattern $A$ and $B$ are the only ones that allow to take out a passenger from $C_I$, so a feasible plan has at least $t$ of those patterns. Once all the passengers are out, (A$\Delta$) ensures that there is no reason to come back to $C_I$. Apply the same reasoning to $C_G$ with $A$ and $\bar B$ to prove the second point. The third point is a simple substraction.
To go to $C_G$ from $C_I$ there is a need for a pattern $A$ but as all the planes are finishing in $C_G$, there is at least $p$ pattern $A$. If there is a pattern $\bar A$ there is a need of another pattern $|A|$, thus proving the fourth point. \hfill $\square$\\

\noindent
{\bf Corollary II}: If a plan does not perform exactly $2t -p$ patterns, it is dominated.\\

\noindent
{\bf Proof}: Straightforward consequence of Property II. \hfill $\square$\\

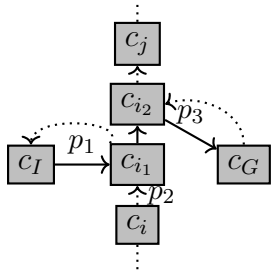
\begin{figure}
\centering
\begin{minipage}{.2\textwidth}
 \begin{tikzpicture}[thick,scale=0.4]
  \node[draw, fill=black!25] (0) at (-3.5,-2) {\large $c_I$};
  \node[draw, fill=black!25] (1) at (0,-2) {\large $c_{i_1}$};
  \node[draw, fill=black!25] (2) at (0,0) {\large $c_{i_2}$};
  \node[draw, fill=black!25] (3) at (0,2) {\large $c_{j}$};
  \node[draw, fill=black!25] (4) at (3.5,-2) {\large $c_G$};
  \node[draw, fill=black!25] (5) at (0,-4) {\large $c_i$};
 
  \draw[thick, ->] (0)--(1) node[midway, above]{$p_1$};
  \draw[thick, ->] (1)--(2) node[midway, right ]{};
  \draw[dotted, ->] (2)--(3) node[midway, above ]{};
  \draw[thick, ->] (2)--(4) node[midway, above ]{$p_3$};
  \draw[dotted, ->] (5)--(1) node[midway, right ]{$p_2$};
  \draw[dotted, ->] (4.north) to [out=90,in=20](2.east) node[midway, right ]{};
  \draw[dotted, ->] (1.north west) to [out=50,in=90](0.north) node[midway, above]{};
  \draw[dotted] (5)--(0,-5.5) node[midway, right ]{};
  \draw[dotted] (3)--(0,3.5) node[midway, right ]{};
 
 \end{tikzpicture}
 \end{minipage}
 \caption{\label{PB} The only non-trivial case where a plan rearrangement is not trivial. One passenger travels through two cities $c_{i_1}$ and $c_{i_2}$ using three planes, respectively $p_1$, $p_2$ and $p_3$. The path taken by the passenger is in bold. By Property II, $p_1$ transports the passenger by performing a pattern $X_1 \in \{ A, B \}$ and $p_2$ a pattern $X_3 \in \{ \bar A, \bar B \}$. We do not assume the pattern performed by $p_2$.}
\end{figure}

By using (A$\Delta$) and given any plan such that a passenger crosses more than one city using two planes, it is easy to find a reorganization of the plan such that it uses only one city and dominates the previous one. However, if a passenger lands in more than one city using at least three planes (one flying full between two central cities), it is not clear whether such a reorganization is possible. This case is illustrated in Figure \ref{PB}. Such a sequence always starts by a pattern $A$ or $B$ and ends by a pattern $\bar A$ or $\bar B$. As the method to reorganize a plan is independent of the original number of cities a passenger goes through, without loss of generality, we will consider the case where a passenger goes through two cities using three planes. 
 
\noindent

Fixing $|A|$ fully determines the cardinal of all patterns and as $|A| \in [p,t]$ we can parameterize the pattern distribution by a single integer $k \in [0, t-p]$. For a given $k$, $\Psi(k)$ denotes the set of elements indicating for each pattern the list of cities. We characterize such a partition w.r.t. a given $k$ by: 

\begin{align*}
  \Psi(k) &= \{ \psi(k) \} \text{~ s.t. ~} \psi(k) \text{~ of the form ~}\\
  \psi(k) & = \left\{\begin{matrix*}[l]
   \;a &:= (a_1, & ..., & \;a_{p+k}) \\ 
   \;\bar a &:= (\bar a_1, & ..., & \;\bar a_{k}) \\
   \;b &:= (b_1, & ..., &\;b_{t-p-k}) \\\
   \bar b &:= (\bar b_1, & ..., &\;\bar b_{t-p-k})\\
  \end{matrix*}\right.
\end{align*} such that any element $e$ of any of the four tuples describes a pattern execution, i.e. $e \in \{1,...,n\}^{|e|}$ with $|e|$ the number of cities involved in the pattern.

For the sake of readability, we denote $\psi(k)$ by $(k, \psi)$ where implicitly, $\psi \in \Psi(k)$. For each couple $(k, \psi)$ we denote by $\mathcal{P}(k, \psi)$ the set of all feasible plans respecting the induced conditions. For a given instance of \MULTIZENO, it is easy to see that $\underset{(k, \psi)}{\bigcup} \mathcal{P}(k, \psi)$ is a partition of the set of feasible plans respecting Property II. In other words, for any feasible plan $p$, there exists an element $(k, \psi)$ such that $p \in \mathcal{P}(k, \psi)$\\

~\\\noindent
For any $k$, $\Psi_0(k)$ denotes the subset of $\Psi(k)$ such that each pattern has a null multiplicity. The elements of the union of $\mathcal{P}(k, \psi_0)$ for any $k$ and $\psi_0 \in \Psi_0(k)$ are the feasible plans using only $2t -p$ cities. Notice that $\mathcal{P}(k, \psi_0)$ may be empty, for instance if a city $b_i$ in $b$ is not present in $\bar b$. In such a case, it is impossible to create a feasible plan respecting the induced constraints. The idea of the demonstration is to show how we can transform any $(k, \psi)$ into a $(k, \psi_0)$ such that there exists $p \in \mathcal{P}(k, \psi_0)$ such that for any $p' \in \mathcal{P}(k, \psi)$, $p \succeq p'$.

The idea is to arbitrarily chose one city for each pattern such that each pattern as a null multiplicity. The only problem is with the pattern $B$ and $\bar B$ that may not have joint city, i.e. one plane will carry a passenger in city and no other plane will come to bring it to $C_G$. Then, we show it is always possible to {\it repair} such a plan, such that the new plan has a lower cost and makespan by construction and uses only $2t-p$ cities.\\

\noindent
{\bf Assumption}: (Symmetry) $\forall i \in [1,n], ~ d_{i} = \bar d_{i}$ \hfill (S)\\

\noindent
{\bf Definition}: (Pattern Reduction) For any pattern with a multiplicity $\theta > 0$ such that it goes through the cities with indices $(i_1,..., i_K)$ (with the first and the last city being $I$ or $G$ depending on the considered pattern), the pattern reduction operation consists in selecting a single city among $(i_2,..., i_{K-1})$ to create a new {\it reduced} pattern of the same type but with multiplicity $0$.\\

\noindent
The reduced pattern has obviously a lower cost as it requires less landings and its duration is lower due to the Assumption $A\Delta$.\\

\noindent
\begin{figure}
\centering
\begin{minipage}{.8\textwidth}
 \begin{tikzpicture}[thick,scale=0.4]
  \node[draw, fill=black!25] (0) at (-3.5,-2) {\large $c_I$};
  \node[draw, fill=black!25] (1) at (0,-2) {\large $c_{i_2}$};
  \node[draw, fill=black!25] (2) at (0,0) {\large $c_{i_3}$};
  \node[draw, fill=black!25] (3) at (0,2) {\large $c_{i_4}$};
  \node[draw, fill=black!25] (4) at (3.5,-2) {\large $c_G$};
  \node[draw, fill=black!25] (5) at (0,-4) {\large $c_{i_1}$};
 
  \draw[thick, ->] (0.south)--(5.west) node[midway, above]{};
  \draw[thick, ->] (1)--(2) node[midway, right ]{};
  \draw[thick, ->] (2)--(3) node[midway, above ]{};
  \draw[thick, ->] (3.east)--(4.north) node[midway, above ]{};
  \draw[thick, ->] (5)--(1) node[midway, right ]{};
 
 \end{tikzpicture}
\hfill $\to$ \hfill
 \begin{tikzpicture}[thick,scale=0.4]
  \node[draw, fill=black!25] (0) at (-3.5,-2) {\large $c_I$};
  \node[draw, fill=black!25] (1) at (0,-2) {\large $c_{i_3}$};
  \node[draw, fill=black!25] (2) at (3.5,-2) {\large $c_G$};
 
  \draw[thick, ->] (0)--(1) node[midway, above]{};
  \draw[thick, ->] (1)--(2) node[midway, above]{};
 
 \end{tikzpicture}
 \end{minipage}
 \caption{\label{pattern_reduction} Pattern reduction for a pattern $A$ such that $|A| = 4$, using city $i_3$.}
\end{figure}
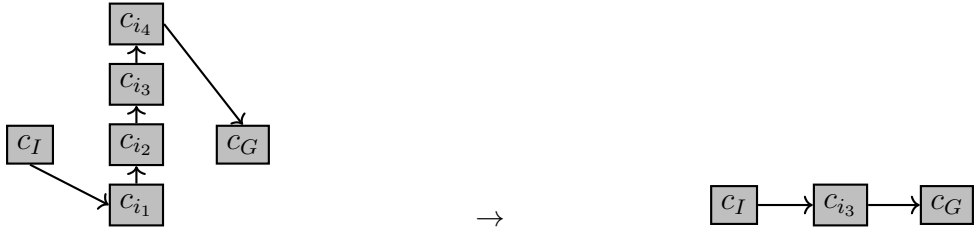

\noindent
{\bf Definition}: (($B\bar B$)-pairing) Consider a pattern $B$ and $\bar B$ in a list of patterns, both of multiplicity $0$ and using respectively the central city $C_i$ and $C_j$. Pairing them consists in using the central city $C^* = \text{argmin} ~ (d_{i} + \bar d_{i}, d_{j} + \bar d_{j})$. \\

\noindent
In the symmetric case, the condition becomes $C^* = \text{argmin} ~ (d_i, d_j)$. Under Assumption $(S)$, the duration of both patterns involved in the pairing is lower or equal to its duration before the pairing.

\noindent
\begin{figure}
\centering
\begin{minipage}{.8\textwidth}
 \begin{tikzpicture}[thick,scale=0.4]
  \node[draw, fill=black!25] (0) at (-3.5,-2) {\large $c_I$};
  \node[draw, fill=black!25] (1) at (0,-2) {\large $c_{i_1}$};
  \node[draw, fill=black!25] (2) at (0,0) {\large $c_{i_2}$};
  \node[draw, fill=black!25] (3) at (3.5,-2) {\large $c_G$};

  \draw[thick, ->] (0)--(1) node[midway, above]{};
  \draw[thick, ->] (2)--(3) node[midway, right ]{};
  \draw[dotted, ->] (3.north) to [out=90,in=20](2.east) node[midway, right ]{};
  \draw[dotted, ->] (1.north west) to [out=50,in=90](0.north) node[midway, above]{};

 \end{tikzpicture}
 \hfill $\to$ \hfill 
 \begin{tikzpicture}[thick,scale=0.4]
  \node[draw, fill=black!25] (0) at (-3.5,-2) {\large $c_I$};
  \node[draw, fill=black!25] (1) at (0,-2) {\large $c_{i_1}$};
  \node[draw, fill=black!25] (2) at (0,0) {\large $c_{i_2}$};
  \node[draw, fill=black!25] (3) at (3.5,-2) {\large $c_G$};

  \draw[thick, ->] (0)--(1) node[midway, above]{};
  \draw[thick, ->] (1)--(3) node[midway, right ]{};
  \draw[dotted, ->] (3.north) to [out=90,in=50](1.north east) node[midway, right ]{};
  \draw[dotted, ->] (1.north west) to [out=50,in=90](0.north) node[midway, above]{};

 \end{tikzpicture}
 \end{minipage}
 \caption{\label{pairing} ($B\bar B$)-pairing assuming $d_{I,i_{j}} + d_{i_{j}, G} < d_{I,i_{j+1}} + d_{i_{j+1}, G}$. }
\end{figure}
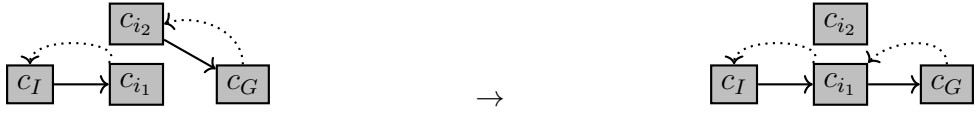
\noindent

Let us consider a plan $p_1$ such that there is at least one pattern (of any type) with a multiplicity higher than 0. We can always apply the following transformation on its list of patterns:

\begin{enumerate}
\item for each pattern with a non-zero multiplicity, do a pattern reduction.
\item for each couple of patterns $B$ and $\bar B$, do a ($B\bar B$)-pairing.
\end{enumerate}
\noindent
It is always possible to pair the patterns since we consider only plans respecting the Property II and in particular the following pattern constraint: $|B| = |\bar B|$.\\

Those steps transform any $(k, \phi)$ into a $(k, \phi_0)$. Indeed, the cardinality of each type of patterns defined by $k$ is conserved but for each single pattern, its multiplicity is now zero. On top of that, each city that appeared in $\phi_0$ also appears in $\phi$ which implies a lower cost of all feasible plans on $(k, \phi_0)$. The set $\mathcal P(k, \phi_0)$ is not empty since we constructed a valid plan.
The duration of each pattern remains the same or is lower than in the original plan due to the reduction. The ($B\bar B$)-pairing also let unchanged the duration for one of the pattern and lower it for the second one involved in the couple due to the symmetric assumption (S). As a result, the transformed plan dominates the original one while using only $2t-p$ cities, thus proving Proposition I under Assumption (A$\Delta$) and Assumption (S). \hfill $\square$\\

In conclusion, we proved that for the Symmetric \MULTIZENO{} problem, Pareto Optimal plans are plans using exactly $2t-p$ central cities. This will allow a constructive algorithm to drastically reduce the search space of feasible and Pareto-optimal plans.

\subsection{Non-Symmetric Clique \MULTIZENO}
\label{sec:non_symmetric}

In this section, we relax the Assumption (S). As a result, the previous method does not work in the particular case a $(B\bar B)$-pairing is not possible. More precisely, it happens when there is no choice of cities to perform a ($B\bar B$)-pairing because for two cities $C_i$ and $C_j$ (resp. for $B$ and $\bar B$) we have $d_{i} < d_{j}$ and  $\bar d_{j} < \bar d_{i}$. In such case, the ($B\bar B$)-pairing on $C_i$ (resp. $C_j$) would increase the duration of the pattern $\bar B$ (resp. $B$) by $2(\bar d_{i} - \bar d_{j})$ (resp. $2(d_{j} - d_{i})$). As a result, without any other change, the transformed plan has a larger total flight duration for at least one plane which may result in a larger makespan. A reorganization of the plan is not trivial such that instead, we will further characterize such situations and propose in Section \ref{sec:adapting} a transformation to get rid of them.\\

We will show that for a plan not to be dominated, the three patterns executed by $p_1$, $p_2$ and $p_3$ as illustrated by Figure \ref{PB} must be $B$, $A$ and $\bar B$. We call it a {\it $B\bar A\bar B$ situation} and in particular, $A$ is of multiplicity exactly equals to $1$.

For each plan, denote by $\mathcal{O}$ (for {\it out}) and $\mathcal{I}$ (for {\it in}) the sets of patterns, respectively to take a passenger from $C_I$ and to bring a passenger to $C_G$.
The set $\mathcal{O}$ contains all patterns $A$ and $B$ while $\mathcal{I}$ contains also all $A$ but also $\bar B$.

As proven previously, in a non-dominated plan, a passenger needs to travel through exactly one pattern from $\mathcal{O}$ and one pattern from $\mathcal{I}$.
Therefore, for each particular passenger, there is the choice between the following couples of patterns:
$$
\begin{matrix}
C1 &=& \text{a single } A \text{ with } \theta = 0 \\
C2 &=& (A, A) \\
C3 &=& (A, \bar B)\\
C4 &=& (B, A)\\
C5 &=& (B, \bar B)\\
\end{matrix}
$$
The multiplicity for each pattern in each couple is defined by
$$
\begin{matrix}
M1 &=& (0) \\
M2 &=& (\theta>0, \theta>0) \\
M3 &=& (\theta>0, \theta \geq 0)\\
M4 &=& (\theta \geq 0, \theta>0)\\
M5 &=& (\theta \geq 0, \theta \geq 0)\\
\end{matrix}
$$
Therefore, for each couple of patterns, we can calculate the number of passengers transported from $C_I$ or to $C_G$, on top of the passengers defined by the couple itself:
$$
\begin{matrix}
T1 &= (0, 0) \\
T2 &= (1, 1) \\
T3 &= (0, 1) \\
T4 &= (1, 0) \\
T5 &= (0, 0) \\
\end{matrix}
$$
For instance, it means that $C3$ implies that another passenger moved from a central city to the destination.
Therefore, it implies that we need to select another couple such that the pattern $\mathcal{I} = A$, i.e. $C2$ or $C4$.

We can deduce that:
$$
\begin{matrix}
C3 & \implies& C2 \vee C4 \\
C4 & \implies& C2 \vee C3 \\
C2 & \implies& C2 \vee C4 + C3 \vee C3
\end{matrix}
$$
As there is a finite number of patterns to pick in a feasible and optimal plan, it is impossible to select $C_2$, $C_3$ or $C_4$.
In other words, we can use only $C_1$ and $C_5$.

The only pattern in $C_1$ has a null multiplicity by definition, such that, if a passenger travels through two central cities, it implies that, not only she does it through $B$ and $\bar B$  but also, there exists a central pattern with multiplicity greater than zero. This pattern cannot be $A$.

Assume such case and that the $(B\bar B$)-pairing is not possible. If the central pattern is $B$, then, another passenger has been moved from $C_I$. Therefore, there is another $B$ somewhere in the plan. In total, the situation implies four patterns $B_1$, $\bar B_1$, $B_2$ and $\bar B_2$ and we assumes $B_1$ and $\bar B_1$ could not be paired with any other pattern. However, by construction, $B_2$ is pairable with $\bar B_1$. Therefore, the central city cannot be $B$.

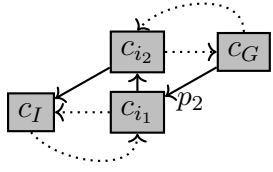
\begin{figure}
\centering
\begin{minipage}{.2\textwidth}
 \begin{tikzpicture}[thick,scale=0.4]
  \node[draw, fill=black!25] (0) at (-3.5,-2) {\large $c_I$};
  \node[draw, fill=black!25] (1) at (0,-2) {\large $c_{i_1}$};
  \node[draw, fill=black!25] (2) at (0,0) {\large $c_{i_2}$};
  \node[draw, fill=black!25] (4) at (3.5,0) {\large $c_G$};
 
  \draw[thick, ->] (2)--(0) node[midway, above]{};
  \draw[dotted, ->] (0.south) to [out=-50,in=-90](1.south) node[midway, above]{};

   \draw[dotted, ->] (1.west) to (0) node[midway, above]{};

  \draw[thick, ->] (1)--(2) node[midway, right ]{};
  \draw[thick, ->] (4)--(1) node[midway, below ]{$p_2$};

  \draw[dotted, ->] (4.north) to [out=90,in=50](2.north) node[midway, above ]{};
  \draw[dotted, ->] (2.east) to (4) node[midway, right ]{};
 
 \end{tikzpicture}
 \end{minipage}
 \caption{\label{BAB_situation} Illustration of a $B \bar A \bar B$ situation. The passenger is carried to and from a central city by a $B$ or $\bar B$ (dashed) and is transported by a $\bar A$ between central cities (bold).}
\end{figure}

By a symmetric reasoning, we conclude that the central city cannot be $\bar B$ and must be $\bar A$ with multiplicity greater than 0 as illustrated by Figure \ref{BAB_situation}.

\noindent

In conclusion, we proved that the proof of the previous section holds except when a $B\bar B$-pairing is not possible. In this case, we proved that non-dominated plans must use a conjunction of three patterns $B\bar A\bar B$ altogether, with $A$ having a multiplicity exactly equal to 1. Therefore, a constructive algorithm can still focus on plans with $2t -p$ central cities with additional care for the particular situation where $B\bar B$-pairing is not possible.

\subsection{General \MULTIZENO}
\label{sec:general_multizeno}

We now consider a connected weighted graph $U = (V,E)$ such that $|E| = n + 2$ and two arbitrary vertices named $I$ and $G$ (with weight $0$), respectively for the initial and the goal cities. A \MULTIZENO\ instance $\Pi$ is defined by the triplet $(U, I, G)$. We denote by $\Lambda$ the set of paths over $U$, and for any path $p\in U$, $|p|$ is the number of cities in the path. 
We define the functions $\phi$ (resp. $\omega$) as follows:

\begin{align}
\forall p \in \Lambda,~ \phi(p) & = \sum_{0 \leq i \leq |p-1|} d_{p_{i,i+1}}\\
\forall p \in \Lambda,~\omega(p) & = \sum_{1 \leq i \leq |p|} c_{p_i}
\end{align}

The function $\phi$ provides the duration, while $\omega$ provides the landing cost of the path. Notice that the first city in a path does not appear in $\omega$ because the initial state is such that the planes are in the initial city.

The following algorithm $f$ allows to transform the general \MULTIZENO\ transport problem defined by $(U, I, G)$ into the original non-symmetric clique problem.

\begin{itemize}
\item For each vertex $i$ find all the paths from $I \to i$ (resp. from $G \to i$) and denote this set $\Lambda_{I \to i}$ (resp. $\Lambda_{i \to G}$).
\item Construct a new graph $\bar U = (\bar V, \bar E)$  such that 
for each $(w_i, e_i) \in \Lambda_{I \to i} \times \Lambda_{i \to G}$, create a vertice of weight $\omega(e_i) + \omega(w_i)$\footnote{The cost of landing in city $C_i$ is counted only once, in the west path.} and an edge $I \to i$ (resp. $i \to G$) of weight $\phi(w_i)$ (resp. $\phi(e_i)$). 
\item For all couple of cities $(i,j) \in \bar V^2$, assign $d_{i,j} = +\infty$.
\end{itemize}

~\\\noindent
{\bf Proposition:} A solution to the transformed problem is a solution to the generic problem.

~\\\noindent
{\bf Proof:} Let $\Pi$ be an instance of the generic problem and $\Pi^*$ the clique instance obtained by the reduction function $f$, i.e. $\Pi^* = f(\Pi)$. We need to show that $p\in P_s(\Pi) \Leftrightarrow p^* = f(p) \in P_s(\Pi^*)$.

~\\\noindent The function $f$ is surjective: $\forall p^* \in \mathcal{P}(\Pi^*)$, $f^{-1}(p^*)$ is the (unique) plan such that we expand every vertex by the associated path in $\Lambda_{I \to i} \times \Lambda_{i \to G}$. For a given $p \in \mathcal{P}(\Pi)$ there exist as many $p^*\in \mathcal{P}(\Pi^*)$ as there are ways of splitting a sequence of cities into two. The application $f^{-1}$ is obviously injective. \\

\noindent
By construction, $\forall p \in \mathcal{P}(\Pi), \forall p^*_1, p^*_2 \in (\text{Im}_f(p))^2, ~  M(p^*_1) = M(p^*_2)$ and $C(p^*_1) = C(p^*_2)$. Furthermore, $\forall p \in \mathcal{P}(\Pi), \forall p^* \in \text{Im}_f(p) 
, ~ M(p) = M(p^*)$ and $C(p) = C(p^*)$ by construction of $f$. Therefore, $f^{-1}$ defines an equivalence relation whose classes are uniquely identified by a $p \in \mathcal{P}(\Pi)$\footnote{They are defined by $p$ and not by $M$ and $C$ since it might exist two plans in $\mathcal{P}(\Pi^*)$ with the same objective vector but a different image by $f^{-1}$.}. 
\noindent
As a result, $f$ defines a bijection from $\mathcal{P}(\Pi)$ to $\mathcal{P}(\frac{\Pi^*}{f^{-1}})$. As for all $p^* \in \mathcal{P}(\frac{\Pi^*}{f^{-1}})$ the objective vectors of $p^*$ and $f^-1(p^*)$ are the same, $p\in P_s(\Pi) \implies p^* = f(p) \in P_s(\Pi^*)$. \hfill $\square$

~\\\noindent
By construction, for any generic instance $\Pi$, the reduced instance $\Pi^*$ satisfies the assumption (A$\Delta$) such that the method described in Section \ref{shortestMK} to identify the Pareto Front directly apply to any instance.

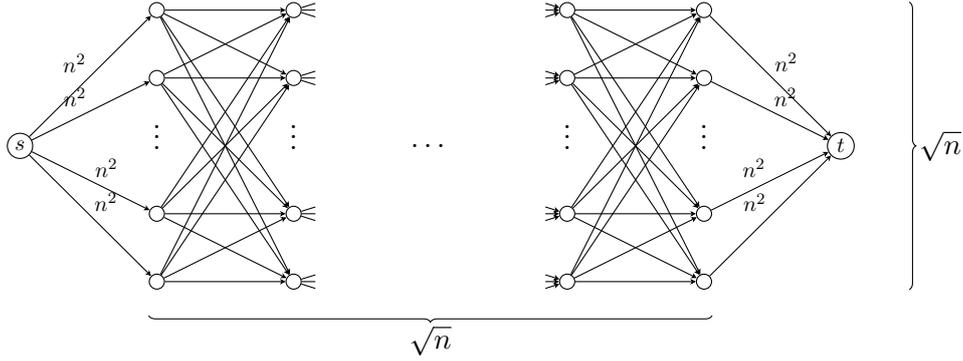
\begin{figure}[tb]
\centering
\begin{tikzpicture}[auto,x=10cm,y=5cm,scale=0.18,
                    vertex/.style={draw,circle,inner sep=2pt},
                    label/.style={scale=0.75,inner sep=2pt}]
  \node[vertex,scale=0.75] (s) at (0,3) {$s$};
  \foreach \x in {1,2,4,5} {%
    \foreach \y in {1,2,4,5} {%
      \node[vertex] (v\x\y) at (\x,\y) {};
    }
    \node (v{\x}3) at (\x,3.25) {$\vdots$};
  }
  \node (v33) at (3,3) {$\dots$};
  \node[vertex,scale=0.75] (t) at (6,3) {$t$};

  \foreach \y in {1,2,4,5} {%
    \path[-{Stealth[width=0.75mm,length=0.75mm]}]
      (s)    edge node[label] {$n^2$} (v1\y)
      (v5\y) edge node[label] {$n^2$} (t)

      \foreach \z in {1,2,4,5} {%
        (v1\y) edge (v2\z)
        (v4\y) edge (v5\z)
      };

      \foreach \z in {-5,0,5} {%
        \path[draw,-]
          (v2\y) -- ($(v2\y.east) + (10mm,\z mm)$);
        \path[draw,{Stealth[width=0.75mm,length=0.75mm]}-]
          (v4\y) -- ($(v4\y.west) + (-10mm,\z mm)$);
      }
  }

  \draw[decoration={brace,mirror},decorate]
    ($(v11.west) + (0,-0.5)$) -- ($(v51.east) + (0,-0.5)$)
      node[midway,swap] {$\sqrt{n}$};
  \draw[decoration={brace,},decorate]
    ($(v55.north) + (1.5,0)$) -- ($(v51.south) + (1.5,0)$)
      node[midway] {$\sqrt{n}$};

\end{tikzpicture}
\caption{\label{fig:super-exp} An example of graph with super-exponential number of s-t paths depending on the number of nodes. The graph is made of $\sqrt n$ layers of $\sqrt n$. Each unlabelled edge has a unit weight. The number of $s-t$ path is then $(\sqrt n)^{\sqrt n}$.}
\end{figure}

~\\\noindent
{\bf On the complexity:} In general the number of cities in $\Pi^*$ is not polynomial in function of $n$, the number of cities in $\Pi$ and thus, solving $\Pi$ through $\Pi^*$ might be challenging w.r.t. the initial complexity of the problem and our algorithm. On top of that, the number of paths between two vertices can be up to super-exponential, as illustrated in Figure \ref{fig:super-exp}, and computing the cardinal of the set of paths is already a $\sharp P$-complete problem \cite{doi:10.1137/0208032}. 

We can notice that in our case, every sub-path of a path in a Pareto-optimal Plan is a non-dominated path itself. Consider a plan $p \in \mathcal{P}(\Pi)$ such that there exists two cities $i$ and $j$  such that the path between those two cities is dominated by another path. It then obvious that the plan $p'$ based on $p$ but using the non-dominated path is at least as good as $p$ because there can not be any $B\bar A\bar B$ situation by construction. Therefore, in non-dominated plans, all paths are non-dominated and all sub-path of a non-dominated path is non-dominated. As a result, it is unnecessary to turn the dominated paths from $\Pi$ into the cities of $\Pi^*$\footnote{It is possible to lose some elements of the Pareto Set, i.e. in the decision space, but the Pareto Frontier will be entirely found.}.

\begin{figure}[tb]
\centering
 \begin{tikzpicture}[thick,scale=1]
  \node[draw, fill=black!25] (0) at (-2,0) {$C_I$};
  \node (1.l) at (0,0.5) {1};
  \node[draw, fill=black!25] (1) at (0,0) {$C_2$};
  \node (2.l) at (-1,-2.3) {$\frac{(n-1)}{2}$};
  \node[draw, fill=black!25] (2) at (-1,-1.5) {$C_1$};
  \node (3.l) at (2,0.5) {1};
  \node[draw, fill=black!25] (3) at (2,0) {$C_4$};
  \node (4.l) at (1,-2.3) {$\frac{(n-3)}{2}$};
  \node[draw, fill=black!25] (4) at (1,-1.5) {$C_3$};
  \node (5.l) at (4,0.5) {1};
  \node[draw, fill=black!25] (5) at (4,0) {$C_{2i}$};
  \node (6.l) at (3,-2.3) {$\frac{(n-(2i-1))}{2}$};
  \node[draw, fill=black!25] (6) at (3,-1.5) {$C_{2i-1}$};
  \node[draw, fill=black!25] (7) at (6,0) {$C_G$};
  \node (8.l) at (5,-2.3) {$\frac{(n-(2i+1))}{2}$};
  \node[draw, fill=black!25] (8) at (5,-1.5) {$C_{2i+1}$};
 
  \draw[thick, ->] (0) -- node[midway, above]{$\frac{n^2}{4}+1$} (1);
  \draw[thick, ->] (0) -- node[left]{1} (2);
  \draw[thick, ->] (2) -- node[left]{1} (1);

  \draw[thick, ->] (1) -- node[midway, above]{$\frac{n^2}{4}+1$}(3);
  \draw[thick, ->] (1) --node[left]{2} (4);
  \draw[thick, ->] (4) --node[left]{2} (3);

  \draw[dotted, ->] (3) -- node[midway, above]{$\frac{n^2}{4}+1$} (5);
  \draw[dotted, ->] (3) --node[left]{$i$} (6);
  \draw[thick, ->] (6) --node[left]{$i$} (5);
  
  \draw[dotted, ->] (5) -- node[midway, above]{$\frac{n^2}{4}+1$} (7);
  \draw[thick, ->] (5) --node[left]{$i+1$} (8);
  \draw[dotted, ->] (8) --node[left]{$i+1$} (7);

\end{tikzpicture}
\caption{\label{fig:hansen} A modified instance of Hansen graph for which the set of non-dominated paths between $C_I$ and $C_G$ is exponential in $n$.} 
\end{figure}
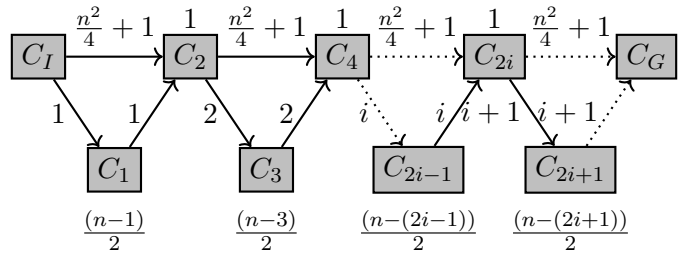

However, even in this case, the number of central cities in $\Pi^*$ is function of the cardinal of the Pareto Set of non-dominated paths for each couple of cities in $\Pi$. In \cite{Hansen1980}, Hansen proposed some pathological instances of bicriterion graphs such that the set of non-dominated paths between two extremes nodes is exponentiel in the size of nodes $n$. We slightly modified the instance to fit our problem as illustred by Figure \ref{fig:hansen}. In this instance, all the paths between $C_I$ and $C_G$ are non-dominated and there are exactly $2^{\frac{n-1}{2}}$ paths.

For computational purposes, we show that if an instance respects an extended version of the triangular inequality, then the number of non-dominated paths is bounded by a polynomial in $n$. This assumption holds, for instance, for planar straight line graph where nodes are associated to points in a Euclidean plane.

~\\\noindent
{\bf Proposition IV:} If for a graph $U$ the number of non-dominated paths is bounded by a polynomial of $n$, the reduction is polynomial of $n$ and the number of cities in $\Pi^*$ is polynomial of $n$.

~\\\noindent
{\bf Proof:} Since any subpath of a non-dominated path must be non-dominated, and the number of non-dominated paths is bounded by a polynomial of $n$, finding the set of non-dominated paths can be done in $n^k$ for a certain $k$. For each non-dominated paths from $C_I$ to $C_G$, the reduction consists in splitting the paths for each city in the paths to create a new city. Let us denote by $D$ the number of non-dominated paths from $C_I$ to $C_G$ and by $L$ the maximal number of central cities of a non-dominated path from $C_I$ to $C_G$. For any non-dominated path from $C_I$ to $C_G$, there are at most $L$ new cities to create. Therefore, the total complexity is bounded by $n^k D L$. \hfill $\square$\\

In conclusion, we know have a method to reduce any general \MULTIZENO~instance to a symmetric \MULTIZENO~instance. Therefore, any solver for the symmetric \MULTIZENO~problem can solve the general \MULTIZENO~problem after transformation. We leave for future work the classification of graphs for which the assumption of Proposition IV holds.

\section{Pareto Optimal Plans}
\label{sec:ppp}

We now focus on finding the Pareto Optimal Plans (PPP) from any list of $2t - p$ cities, that is to say from the elements ($k, \psi_0)$, for $k \in [0, t - p]$ and $\psi_0 \in \Psi_0(k)$ for the symmetric and {\bf non-symmetric} case, that is, for the latter, we assume that there is no $B\bar A \bar B$ situations for now.\\

\subsection{Definitions}
\label{sec:ppp_def}

\noindent
{\bf PPPs and Admissible PPPs}: A Possibly Pareto-optimal Plan (PPP) is defined by 3 tuples, namely $a \in \{1,...,n\}^{k+p}$ for cities involved in a pattern $A$, $\bar a \in \{1,...,n\}^{k}$ for cities involved in a pattern $\bar A$, and $b \in \{1,...,n\}^{t-p-k}$ for the cities involved in $B$ and $\bar B$.\\

\noindent
Nevertheless, $a$, $\bar a$ and $b$ do not hold any information about which plane will land in a particular city. This is the reason why there exists many feasible schedules, i.e., schedules that actually are feasible plans for $p$ planes\footnote{Most of them are probably not Pareto-optimal, but w.r.t. Using Proposition I as hypothesis, any schedule resulting from a larger tuple $a$ or $\bar a$ or $b$ would be Pareto-dominated.} using the corresponding $4t-2p$ edges. There are at most $n^{(2t-p)}$ possible PPP but it is clear that the set of PPPs contains many redundancies, that can easily be removed by ordering the indices\\

\noindent
{\bf Definition}: (Admissible PPP) An {\it admissible PPP} is an element of $A \times \bar A \times B$, where $A = \{a \in [1,n]^{k+p} ; \forall i \in [1,k+p], d_{a_i} \geq d_{a_{i+1}}\}$, $\bar A = \{\bar a \in [1,n]^{k} ; \forall i \in [1,k],  d_{{\bar a}_i} \geq d_{{\bar a}_{i+1}}\}$ and $B = \{b \in [1,n]^{t - p - k} ; \forall i \in [1,t-p-k], d_{b_i} \geq d_{b_{i+1}}\}$.\\

\noindent
{\bf Number of admissible PPPs}: Let $K^{m}_{k}$ be the set of $k$-multicombinations (or multi-subset of size $k$) with elements in a set of size $m$. The cardinality of $K^{m}_{k}$ is $\Gamma^{m}_{k} = {m+k-1 \choose k}$. As $A$ is in bijection with $K^{n}_{k+p}$, $\bar A$ with $K^{n}_{k}$ and $B$ with $K^{n}_{t - p - k}$, the number of PPP is $(t - p)\Gamma^{n}_{k+p} \Gamma^{n}_{k} \Gamma^{n}_{t - p - k}$, i.e., $(t - p){n+k+p-1 \choose k + p}{n + k - 1 \choose k}{n + t - p - k - 1 \choose t - p - k}$.\\

\noindent
{\bf Cost of a PPP}: Given the PPP $\psi_0 = (a, \bar a, b) \in A \times \bar A \times B$, the cost of {\bf any} plan using only the cities in $a$, $\bar a$ and $b$ is uniquely defined by $\text{Cost}(\psi_0) = \underset{{a_i \in a}}{\sum} c_{a_i} + \underset{{{\bar a}_i \in \bar a}}{\sum} c_{\bar a_i} + 2\underset{{b_i \in b}}{\sum} c_{b_i}$.\\

\noindent
{\bf Makespan of a PPP}: The makespan of a PPP is thus that of the shortest schedule that uses its $4t-2p$ edges in a feasible way. Trivial upper and lower bounds for the shortest makespan of a PPP $\psi_0$ are respectively $M_S(\psi_0)$, the makespan of the sequential plan (i.e., that of the plan for a single plane that would carry all persons one by one), and $M_L(\psi_0)$, the makespan of the perfect plan where none of the $p$ planes would ever stay idle and the length can be perfectly shared between the planes. These bounds are useful to prune the set of PPPs:

\begin{align*}
M_S(\psi_0) & = \underset{{a_i \in a}}{\sum} (d_{a_i} + \bar d_{a_i}) + \underset{{{\bar a}_i \in \bar a}}{\sum} (\bar d_{\bar a_i} + d_{\bar a_i}) + \underset{{b_i \in b}}{\sum} (d_{b_i} + \bar d_{b_i})\\ 
M_L(\psi_0) & = \frac{M_S(\psi_0)}{p}
\end{align*}

\noindent
{\bf $\Psi$-domination}: Given two PPP $(k, \psi_0)$ and $(k', \psi_0')$, $\psi_0$ {\it $\Psi$-dominates} $\psi_0'$ if $M_S(\psi_0) \leq M_L(\psi_0')$ and $Cost(\psi_0) \leq Cost(\psi_0')$.\\

\noindent
$\Psi$-domination is different from the standard domination as it occurs in a different space: we do not compare plans but the PPP structures which can individually lead to several plans. Note that if $\psi_0 \overset{\Psi}{\succeq} \psi_0'$, there is no need to compute the shortest makespan for $\psi_0'$ because any plan in $\mathcal{P}(k', \psi_0')$ is dominated by any plan in $\mathcal{P}(k, \psi_0)$.

\subsection{Computing the Shortest Makespan and Constructing the Plan}
\label{shortestMK}

The method to compute the optimal makespan for a particular PPP $\psi_0$ is broken down into four steps. All the steps are performed greedily allowing a resolution per PPP in linear time in function of the size of $\psi_0$. After detailing these steps, we will give a constructive proof that the obtained makespan is optimal. 


If a non-symmetric instance is highly imbalanced in durations, while performing a $B\bar B$, there is a chance that the plane performing $\bar B$ will have to wait. It implies that its makespan is not the sum of the durations of its patterns but the moment the passenger arrives in the central city, plus the remaining duration of its track. For this reason, we denote by $T_i$ the moment passenger $i$ arrives in the central city she goes through.

\begin{enumerate}
\item For each city $i$ in $b$, greedily distribute by descending order of ${d}_i + {\bar d}_i$ the duration $2 \max ~ ({d}_i, {\bar d}_i)$ among the planes. If the ${d}_i > {\bar d}_i$ add $C_G \to C_i\to C_G$ to the sequence of the plane, otherwise add $C_I \to C_i\to C_I$. If there is already a sequence from $C_I$ (resp. $C_G$) add the new one at the right side (resp. left side) of the existing ones.
\item Greedily distribute the $p$-largest elements of $a$ among the $p$ planes. Note that each plane must receive one duration due to the fact that each plane should finish in $C_G$. For each plane, the sub-sequence $C_I \to C_i \to C_G$ is to be added between the western and eastern parts of the pattern induced by $b$ distributed in the first step.
\item While it remains some elements in $a$ and $\bar a$, select the plane with the minimal duration and add the largest element of $a$ or $\bar a$ depending the previous element it received ($a$ if $\bar a$, and vice versa). For an element of $a$ (resp. $\bar a$), add $C_I \to C_i \to C_G$ (resp. $C_G \to C_i \to C_I$) right before the sequence added during the second step.
\item For each city $i$ in $b$, greedily distribute by descending order of ${d}_i + {\bar d}_i$ the duration $2 \min ~ ({d}_i, {\bar d}_i)$ among the planes. The rules to add the sub-sequences are the same as in the first step. If ${d}_i < {\bar d}_i$, assign to the plane the makespace $max(D(p) + \bar d_i), T_i) + \bar d_i$ where $D(p)$ is the duration of the partial track up to the moment the plane arrives in $C_i$.
\end{enumerate}
The optimal makespan for the given PPP is the longest duration among the $p$ planes.\\

~\\\noindent
{\bf Proposition}: For a given PPP $\psi_0$ and $\beta_{\text{set}}$, the algorithm returns the optimal makespan.\\

\noindent
{\bf Proof for the symmetric case}: The incompressible time to transport all passengers, according to a given PPP is $T(\psi_0) = 2\underset{i\in b}{\sum} (d_{i} + \bar d_{i}) + \underset{i\in \{ a, \bar a \}}{\sum} (d_{i} + \bar d_{i})$. A theoretical optimal plan with this pattern repartition is a plan without any waiting point for any plane.
The above algorithm gives the optimal distribution of the set of times into $p$. Then, if a plan can be constructed with such a makespan, it is optimal for the PPP.
As it constructs such plan, we can conclude that the algorithm is optimal for the PPP, thus proving Proposition~I.\hfill $\square$\\

\noindent 
{\bf Proof for the non-symmetric case}: The above algorithm gives the optimal distribution of the set of times into $p$ and provides a plan that minimizes the waiting time by starting the plan by the patterns that could lead to a waiting time.\hfill $\square$\\

\noindent
{\bf Complexity}: For a given instance, the size of any PPP is $2t - p$. Finding the best makespan for a PPP is linear in $2t - p$. As a result, the complexity to solve an instance is given by $(t - p){n+k+p-1 \choose k + p}{n + k - 1 \choose k}{n + t - p - k - 1 \choose t - p - k} \mathcal{O}(2t - p)$.

\subsection{Adapting the method to $B\bar A\bar B$ situations}
\label{sec:adapting}

A PPP cannot be a structure in which there are $B\bar A\bar B$ situations. Otherwise, there might exist plans using the cities of such PPP, with an additional city for pattern $A$, such that the plan is non-dominated by any plan using only the $2t-p$ cities of the PPP. However, the algorithm is still optimal even if the duration of flights does not only depends on the city but also on the type of patterns. For instance, if we could arbitrarily decide that a pattern $A$ going through $C_i$ has a larger duration than the counterpart $\bar A$ using the same city. In other words, there would be $d_i^X$ and $\bar d_i^X$ for any $X \in \{A, \bar A, B, \bar B \}$.
Using this idea, we can transform any instance with some $B\bar A\bar B$ situations into an instance without any, such that the algorithm A1 is optimal. 

\begin{figure}
\centering
\begin{minipage}{.2\textwidth}
 \begin{tikzpicture}[thick,scale=0.4]
  \node[draw, fill=black!25] (0) at (-3.5,-2) {\large $c_I$};
  \node[draw, fill=black!25] (1) at (0,-2) {\large $c_{k}$};
  \node[draw, fill=black!25] (4) at (3.5,-2) {\large $c_G$};
 
  \draw[thick, ->] (1)--(0) node[midway, above]{};
  \draw[dotted, ->] (0.south) to [out=-50,in=-90](1.south) node[midway, above]{};

   \draw[dotted, ->] (1.west) to [out=50,in=90](0) node[midway, above]{};
  \draw[thick, ->] (4)--(1) node[midway, above ]{$p_2$};

  \draw[dotted, ->] (4.north) to [out=90,in=50](1.north) node[midway, above ]{};
  \draw[dotted, ->] (1.east) to [out=-50,in=-90](4) node[midway, right ]{};
 
 \end{tikzpicture}
 \end{minipage}
 \caption{\label{BAB_transformation} The BAB situation is transformed into a regular $B\bar B$-pairing through a new city and a $\bar A$ with a null multiplicity. The cost of the new city $C_k$ is $c_i$ for $B$, $c_j$ for $\bar B$ and $c_i + c_j$ for $\bar A$ or $A$.} 
\end{figure}
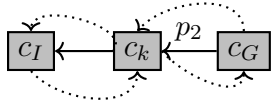

Consider an instance $\Pi$ of the non-symmetric \MULTIZENO~problem. For each potential $B\bar A\bar B$ situation going through $C_i$ and $C_j$, add a new city $C_k$ such that:
\begin{align}
d_k^B = 2d_i \\
d_k^{\bar B} = 2d_j \\
d_k^A = \min(d_i + \bar d_i, d_j + \bar d_j) \\
d_k^{\bar A} = \bar d_i + d_{ij} + d_j
\end{align}
This transformation is illustrated by Figure \ref{BAB_transformation}. The transformation has $\mathcal{O}(n^2)$ complexity and the resulting instance $\Pi^*$ is a non-symmetric \MULTIZENO~without any $B\bar A\bar B$ situation. More precisely, if there is a plan using a $B\bar A\bar B$ that is optimal in $\Pi$, then it is also optimal in $\Pi^*$, but there exists a plan with the same makespan and cost without any $B\bar A\bar B$ situation in $\Pi^*$.

\section{\ZENOSOLVER}
\label{sec:zenosolver}
\ZENOSOLVER\ is a \C++ software dedicated to generate and exactly solve \MULTIZENO\ instances. \ZENOSOLVER\ computes the true Pareto Front using the algorithm described in Section \ref{sec:ppp}. 
It outputs the corresponding \texttt{PDDL} file\footnote{Planning Domain Definition Language~\cite{fox2003pddl2}, almost universally used in the AI Planning to describe domains and instances.}, that can be directly used by most AI planners.

We implemented two versions of the algorithms to iterate over the set of PPP, namely the {\it classic}, by reference to our previous work \cite{quemy:hal-01109777}, and the {\it no-duplicate} version.
Algorithms \ref{classic} and \ref{no_duplicate} present a high-level view of both versions, respectively for classic and no-duplicate. 

\subsection{Classic and no-duplicate}

In the {\bf classic version} (shown on Algorithm~\ref{classic}), we iterate over the set of $t$ tuples that represents the cities involved in patterns going eastward, and then over the set of $t-p$ patterns representing the cities involved in patterns going westward. For each couple of tuple $(e, w)$, we compute the powerset of the intersection. This gives all the possibilities for a $B\bar B$-pairing based on the PPP. Finally, we iterate over this powerset and compute the lower makespan for each triplet $(e \setminus \beta, w \setminus \beta, \beta)$. This version is simple to understand and generates PPP in an approximately increasing order of cost which allows for efficient pruning. However, the method still generate a set of duplicates that grows exponentially with the number of passengers. The duplicates appears both in the space of PPP and the set of plans (because some patterns $A$ and $\bar A$ can sometimes be swapped).

\algrenewcommand\algorithmicindent{1.0em}%
\begin{algorithm}
  \caption{Classic version of \ZENOSOLVER}\label{classic}
  \begin{algorithmic}[1]
    \Procedure{Solver}{$n, t, p, d, \bar d, D, c$}
        \For{$e \in K^{n}_{t}$}
          \For{$w \in K^{n}_{t-p}$}
            \State $C \gets \text{cost}(e, w)$
            \State $B \gets \mathbb{P}(e \cap w)$
            \For{$\beta \in B$}
              \State $M \gets \text{lowestMakespan}(e \setminus \beta, w \setminus \beta, \beta)$
            \EndFor
        \EndFor
      \EndFor
    \EndProcedure
  \end{algorithmic}
\end{algorithm}

The {\bf no-duplicate version} adopts a different view on the construction of PPP. The main loop iterates over the set of tuples $u$ of size $2t-p$.

Then, we generate all possible subsets of $p$ elements, without duplicate. This implements the constraint that each of the $p$ planes will perform a pattern $A$. Let denotes by $P$ this set. For each $m \in P$, it remains a tuple $v = u \setminus m$ of size $2(t-p)$. With $v$, we apply the same method as the classic algorithm: we compute the set of cities possibly involved in a $B\bar B$-pairing and generate its powerset. Iterating over the powerset, we compute the lowest makespan for each triplet $(m, v, \beta)$. 

At first sight, the two algorithms are similar, except that the no-duplicate version does ``block'' the firsts $p$ occurrences of pattern $A$. However, this difference decreases by an exponential factor the computation time in several ways: 1) there is no possible duplicate which decreases the computation time with $n$ by comparison to the classic version, 2) the powerset for the possible $B\bar B$-pairing is done on a smaller set, 3) increasing $p$ decreases the cardinality of $v$. Also, as there is only one main loop, it makes it easier to implement efficient parallelism. The drawback is that the PPP are no longer generated in an approximately increasing order of cost.

The {\em classic} version seems more efficient in terms of effective computational time on problems showing a small number of passengers, while the {\em no-duplicate} one is faster with a growing number of passengers or a compromise between cities and passengers. Also, all other things being equal, the no-duplicate version becomes faster when $p$ increases while the computational time for the classic version remains unchanged. See Section~\ref{sec:perf} for further details.

\algrenewcommand\algorithmicindent{1.0em}%
\begin{algorithm}
  \caption{No-duplicate version of \ZENOSOLVER}\label{no_duplicate}
  \begin{algorithmic}[1]
    \Procedure{Solver}{$n, t, p, d, \bar d, D, c$}
      \For{$u \in K^{n}_{2t - p}$}
        \State $C \gets \text{cost}(e, w)$
        \State $(M, \#c) \gets \text{mapping (cities} \to \text{number of cities in } u$)
        \For{$v \in K^{M}_{2(t - p)}$}
          \State $\bar v \gets \text{determinePossible}B\bar B\text{-pairing}(v)$ 
          \State $B \gets \text{powerset}(\bar v)$
            \For{$\beta \in B$}
              \State $M \gets \text{lowestMakespan}(u, v \setminus \beta, \beta)$
            \EndFor
        \EndFor
      \EndFor
    \EndProcedure
  \end{algorithmic}
\end{algorithm}

Both algorithms are based on two costly operations: 1) generating all multicombinations of $k$ elements among $n$ elements with repetitions, and 2) generating the powerset of a given set of elements. For 1), the implementation follows the one proposed in \cite{Knuth}. Regarding 2), we used the Same Number of One Bit (Soob) technique \cite{beeler1972hakmem} that operates bitwise by noticing that equal cardinality subsets have the same number of one bits.

Using the $\Psi$-domination, \ZENOSOLVER\ implements a pruning method that checks if the current PPP is dominated by any other PPP already stored. As noted, the optimal makespan is lower or equal than the upper bound $M_S$, leading to an efficient pruning. Indeed, as PPPs are generated in an approximated increasing order \cite{Knuth}, this avoids iterating over the whole set to check the domination criterion.

Determining if the current PPP is dominated has complexity $O(h)$ where $h$ is the number of different total achievable costs. An obvious upper-bound for $h$ is given by $(2t-p)(\max_i(c_i) - \min_i(c_i))$. However, in practice, $S$ seems to have the same order of magnitude than the exact Pareto Front. In addition, $S$ is the only structure kept in memory, thus, from this point of view, \ZENOSOLVER\ turns out to be near-optimal regarding the memory usage (see Table \ref{table:i}). \\

\subsection{Handling the non-symmetric instances}

To handle non-symmetric instances, we need the following additional two steps:
\begin{enumerate}
\item We modified the algorithm such that we could specify a different duration and cost for each pattern based on a city. In practice, it does not change anything to the optimality of the algorithm because we only consider the total duration of a pattern rather than individual flights within the pattern.
\item We added an additional preprocessing step prior using the algorithm. The preprocessing step consists in determining the possible $B\bar A\bar B$ situations in the given instance. For each $B\bar A\bar B$ situations, we add a new city as described in Section~\ref{sec:adapting}
\end{enumerate}

\subsection{Empirical Performances}\label{sec:perf}

All experiments have been performed using a VM running Ubuntu, equipped with a 12 cores i7-9750H CPU @ 2.60 GHz, 64 Gb RAM and a NVMe SSD.

\begin{figure}[bt]
  \centering
      \includegraphics[width=0.50\textwidth]{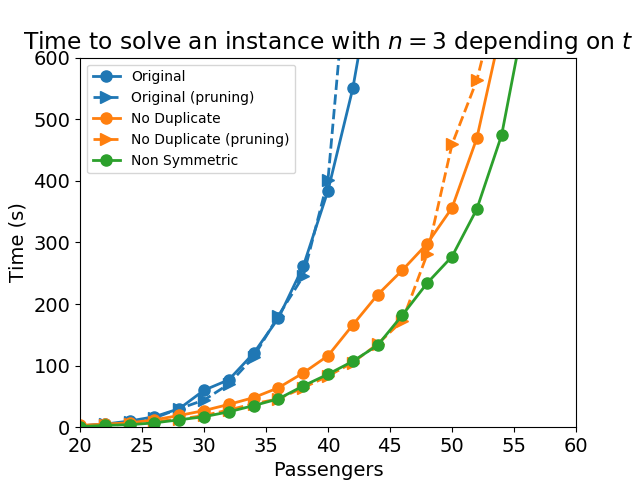}\hfill
      \includegraphics[width=0.50\textwidth]{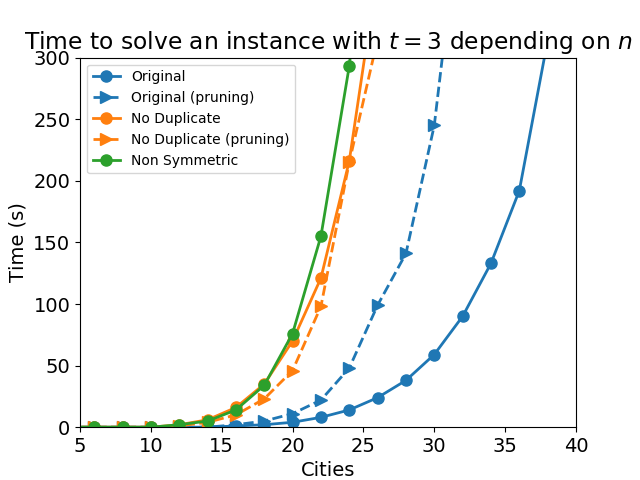}
 \caption{\label{speed1} Time function of $t$ (left) or $n$ (right) for $d_i = \bar d_i = c_i = i$.}
\end{figure}

In Figure \ref{speed1}, we report the time to solve an instance with $d_i = \bar d_i = c_i = i$. On the left, we fixed $n=3$ and on the right, we fixed $t=3$. For both both, the number of planes has been fixed to 2. Three versions are displayed: original, no-duplicate and non-symmetric which is a no-duplicate version that takes into account the $B\bar A\bar B$ situations. There is no pruning for the last version because the $\Psi$-domination introduced in Section \ref{sec:ppp_def} does not hold for the non-symmetric version in case of $B\bar A\bar B$ situations due to the possibility of waiting times. 

As expected, all curves are exponential in their respective parameters. The no-duplicate version allows to solve similar problems about twice as fast as the classic version when the number of passengers increases. Conversely, the classic version provides a similar speed-up when the number of cities increases.

 However, when both $n$ and $t$ grow together the no-duplicate version is clearly better, even with the overhead implied by dealing with the $B\bar A\bar B$ situations. This is clear by looking at Table \ref{table:i} and  \ref{table:i_2} that reports several metrics, for both versions, when $n$ and $t$ increases simultaneously. The number of generated elements is always one to three order of magnitude lower compared to the classic version. Similarly, the call number to the costly routine lowestMakespan is one order of magnitude lower. Interestingly, the intrinsic cost of the no-duplicate version, with the overhead to handle $B\bar A\bar B$ situations is compensated from $n=t=10$.

\begin {table}[bt]
\centering
\caption{\label{table:i} Increasing simultaneously $n$ and $t$ with $d_i=\bar d_i = c_i = i$ for the classic version.}
\begin{tabular}{|c|c|c|c|c|c|c|c|}
  \hline
  n & t & p & Iterations & lowestMakespan calls & $S$ Size & Front Size & Time\\
  \hline
  3 & 3 & 2 & 30 & 33 & 9 & 5 & 0ms\\
  4 & 4 & 2 & 350 & 408 & 19 & 10 & 0ms\\
  5 & 5 & 2 & 4410 & 6387 & 33 & 17 & 3ms\\
  6 & 6 & 2 & $58 \times 10^{3}$ & $10 \times 10^{4}$ & 51 & 26 & 79ms\\
  7 & 7 & 2 & $79 \times 10^4$ & $19 \times 10^{5}$ & 73 & 37 & 1657ms\\
  8 & 8 & 2 & $11 \times 10^6$ & $34 \times 10^{6}$ & 99 & 50 & 31.968s\\
  9 & 9 & 2 & $15 \times 10^7$ & $63 \times 10^{7}$ & 129 & 65 & 703.141s \\
  10 & 10 & 2 & $22 \times 10^8$ & $11 \times 10^{9}$ & 163 & 82 & 4:33h \\
  \hline
\end{tabular}
\end{table}

\begin {table}[bt]
\centering
\caption{\label{table:i_2} Increasing simultaneously $n$ and $t$ with $d_i=\bar d_i = c_i = i$ for the no-duplicate version.}
\begin{tabular}{|c|c|c|c|c|c|c|c|}
  \hline
  n & t & p & Iterations & lowestMakespan calls & $S$ Size & Front Size & Time\\
  \hline
3 & 3 & 2 & 15 & 41 & 9 & 5 & 0ms\\
4 & 4 & 2 & 84 & 454 & 19 & 10 & 0ms \\
5 & 5 & 2 & 495 & 6299 & 33 & 17 & 13ms \\
6 & 6 & 2 & 3003 & $83 \times 10^{3}$ & 51 & 26 & 209ms \\
7 & 7 & 2 & $18 \times 10^{3}$ & $10 \times 10^{5}$ & 73 & 37 & 3.006s \\
8 & 8 & 2 & $11 \times 10^{4}$ & $14 \times 10^{6}$ & 99 & 50 & 37.584s \\
9 & 9 & 2 & $73 \times 10^{4}$ & $17 \times 10^{7}$ & 129 & 65 & 505.111s \\
10 & 10 & 2 & $46 \times 10^{5}$ & $21 \times 10^{8}$ & 163 & 82 & 1:55h \\
  \hline
\end{tabular}
\end{table}

\newpage

\subsection{Examples of Instances}

\begin{figure}[bt]
  \centering
      \includegraphics[width=0.50\textwidth]{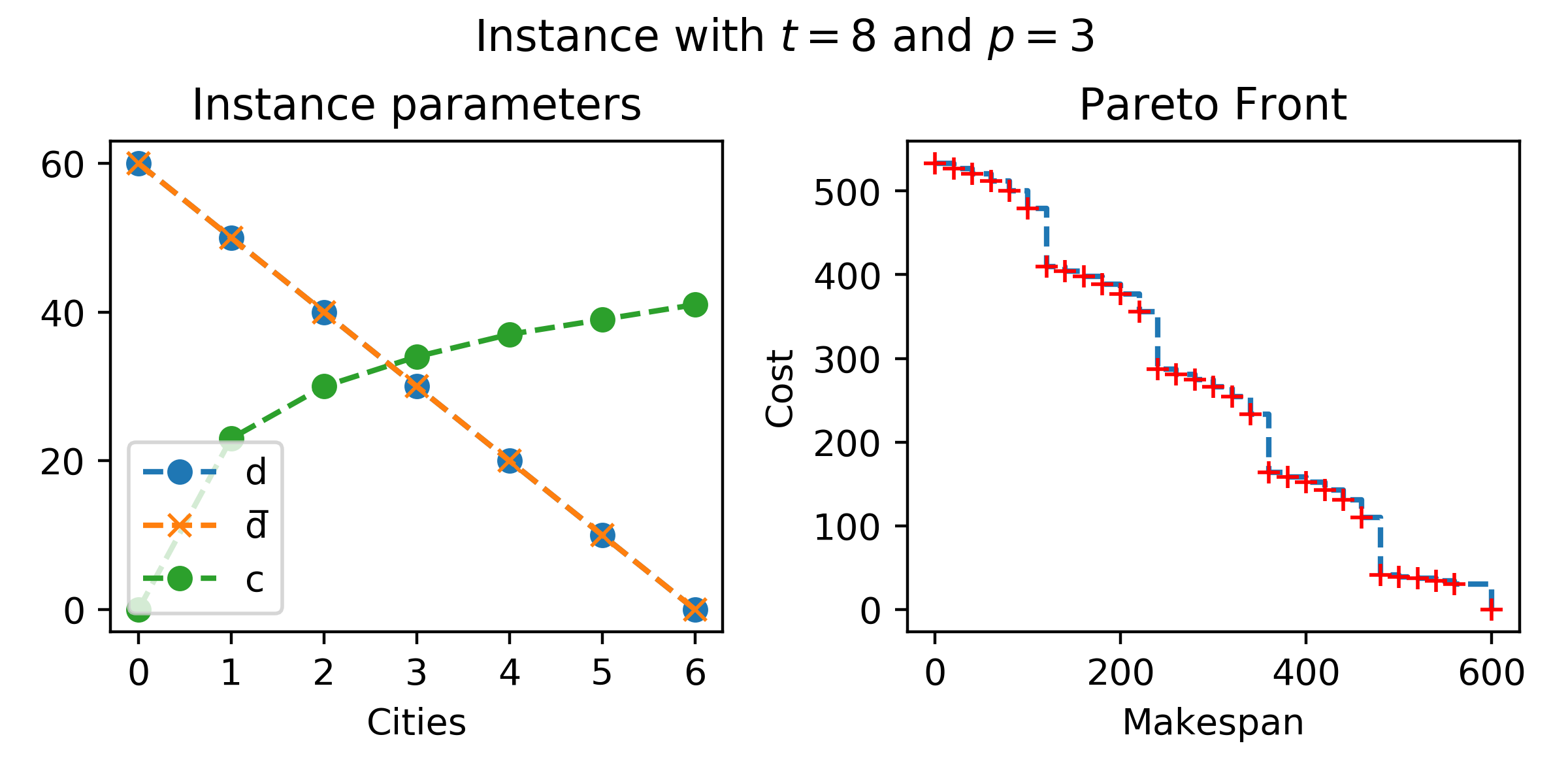}\hfill
      \includegraphics[width=0.50\textwidth]{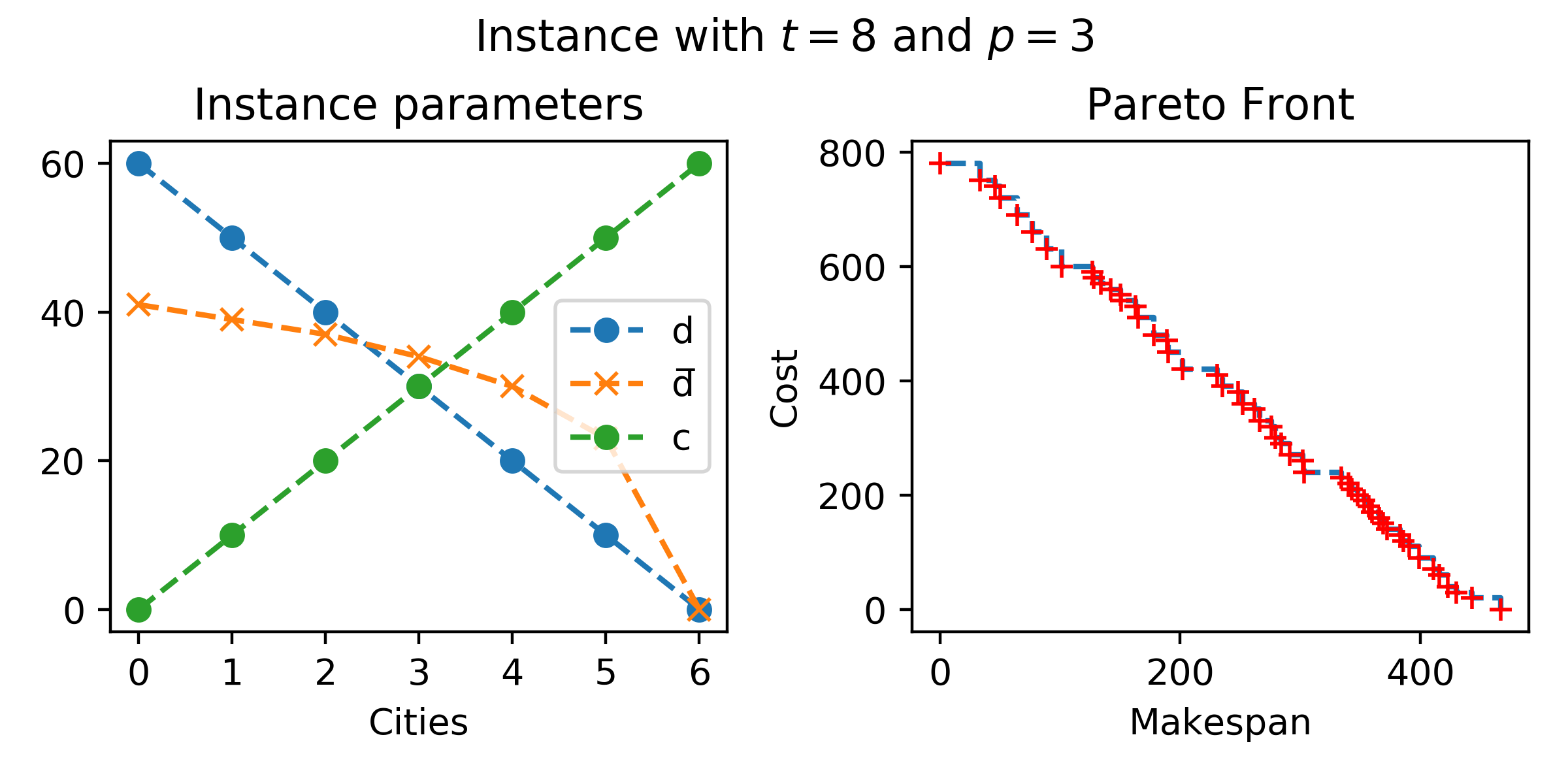}\\
      \includegraphics[width=0.50\textwidth]{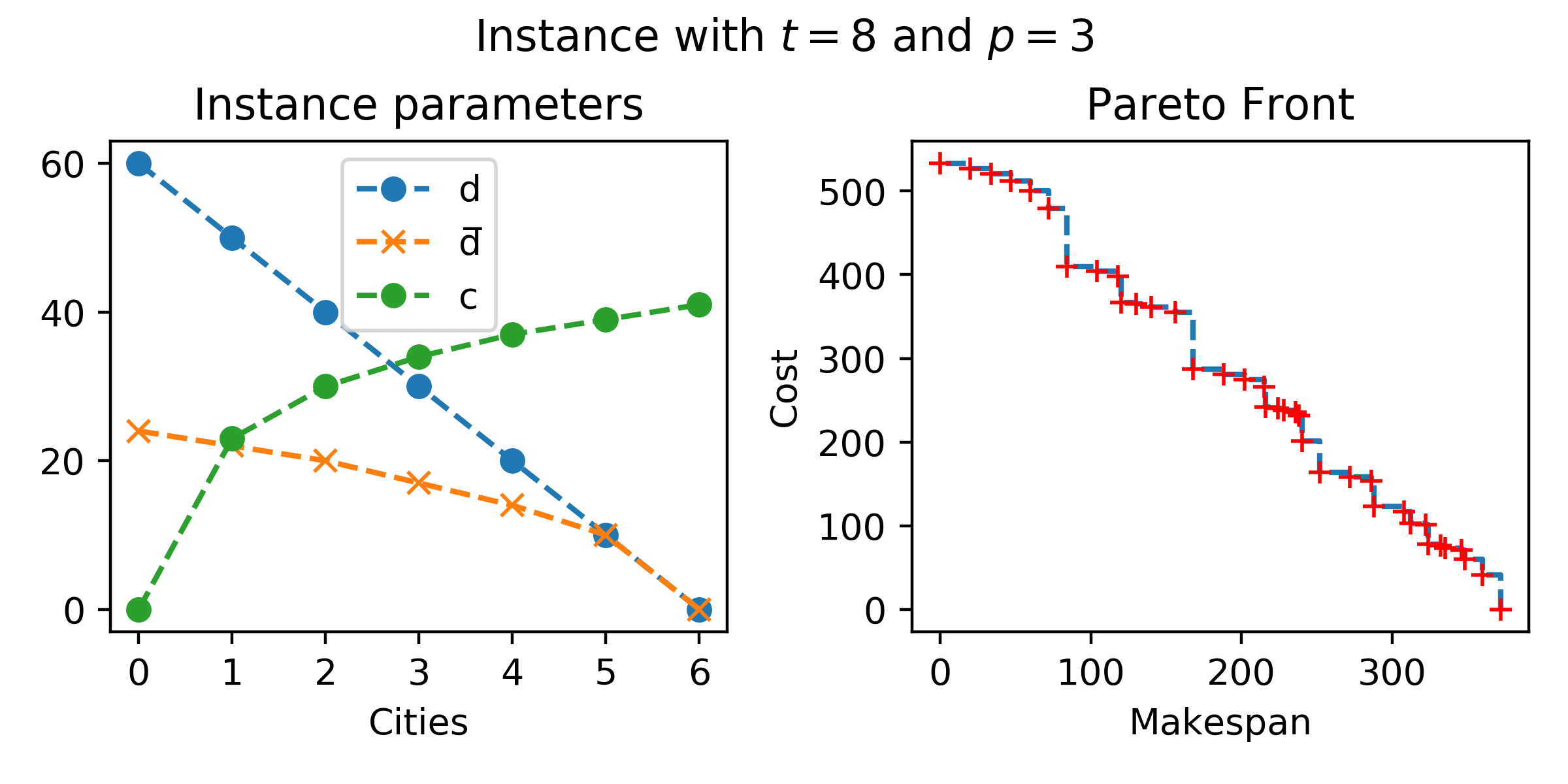}\hfill
      \includegraphics[width=0.50\textwidth]{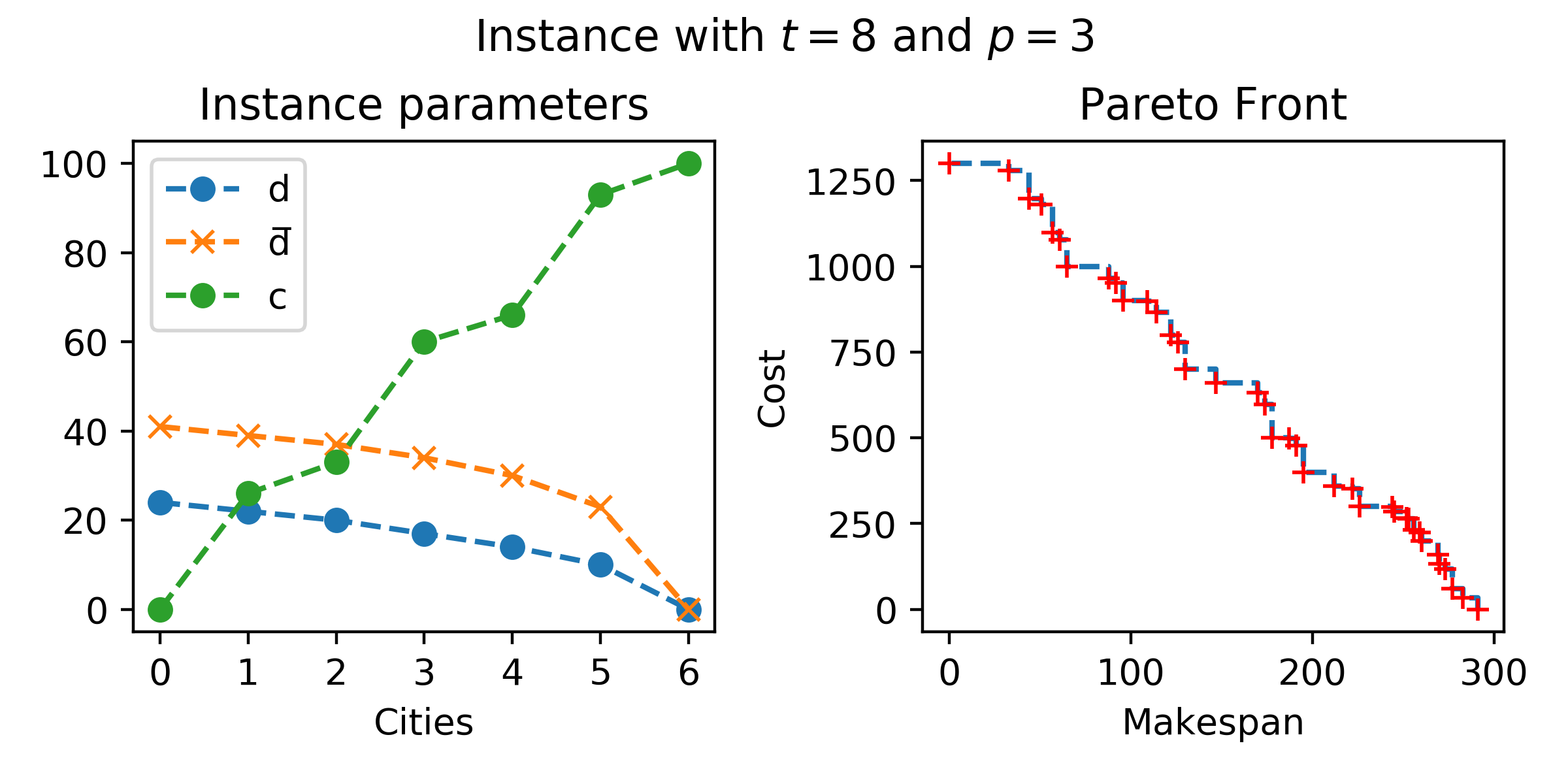}
 \caption{\label{instances} Different instances with $n = 7$, $t=8$ and $p=3$ and various $d$, $\bar d$ and $c$.}
\end{figure}

In Figure \ref{instances}, we display some examples of the variety of Pareto Fronts that it is possible to obtain by modifying the functions to generate $d$, $\bar d$ and $c$. The top left picture shows regular patterns with a uniform disposition of points. This instance is obtained with a cost $c_i = log(i + 1)$. The top right figure is obtained by using $\bar d_i = i$ and also displays some pattern but with a non-uniform point distribution over the front. By using slightly more complex combination of generators, it is possible to obtain non-regular fronts with non-uniform distribution such as the bottom two figures: on the left, $\bar d_i = \sqrt(i)$ and $c_i = log(i + 1)$, while on the right, $d_i = \sqrt(i)$, $\bar d_i = log(i + 1)$ and $c_i = \frac{5}{3} x + x \text{ mod } 2$.

\section{Application: OpenFlight Data}
\label{sec:application}

To generate benchmarks with real data, we used the list of airports and routes from OpenFlight Database\footnote{\url{https://openflights.org/data.html}}. We filtered to keep only the 50 largest airports with regards to the number of passengers per year. We selected as initial airport and goal airport, the largest and second largest airport, namely Hartsfield Jackson Atlanta International Airport (ATL) and Beijing Capital International Airport (PEK). We then filtered to keep only the existing routes between the remaining airports.

For each route, we calculated the spherical distance between the two airports using Haversine formula. This distance is used for the makespan.
The cost of landing in an airport has been defined as follows: for a given airport $C_i$, 1) compute the spherical distances $d_{\text{ATL},i}$ and $d_{i,\text{PEK}}$ between the airport and respectively, ATL and PEK; 2) assign the inverse of the average distance i.e. $c_i = \frac{2}{d_{\text{ATL},i} + d_{i,\text{PEK}}}$.

Then, we generated all simple paths from ATL to PEK between the remaining airports using the existing routes with a maximal path length of 4 cities. We filtered to keep only the Pareto efficient paths.

To generate a symmetric version of \MULTIZENO, we used the reduction presented in Section \ref{sec:general_multizeno} on the remaining simple path between ATL and PEK.

The final symmetric instance has 15 central cities that corresponds to 15 different non-dominated paths from ATL to PEK using a total of 12 airports. Some paths uses only one intermediate airport (e.g. \texttt{ATL -> DXB -> PEK}), while some uses two (e.g. \texttt{ATL -> LAS -> SFO -> PEK}) or three (\texttt{ATL -> DFW -> LAS -> SFO -> PEK}).

\begin{figure}[H]
  \centering
      \includegraphics[width=0.65\textwidth]{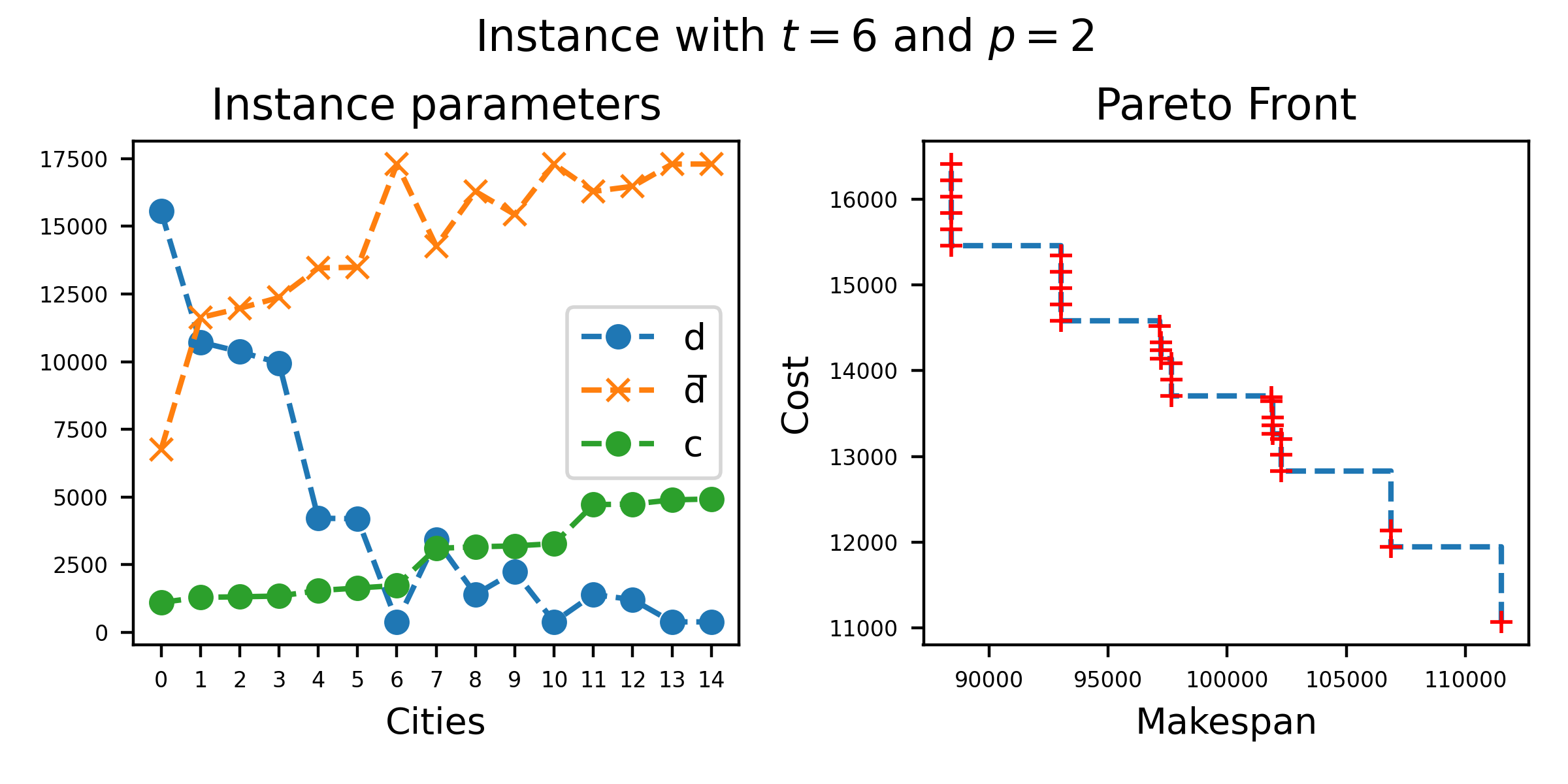}\hfill
      \includegraphics[width=0.65\textwidth]{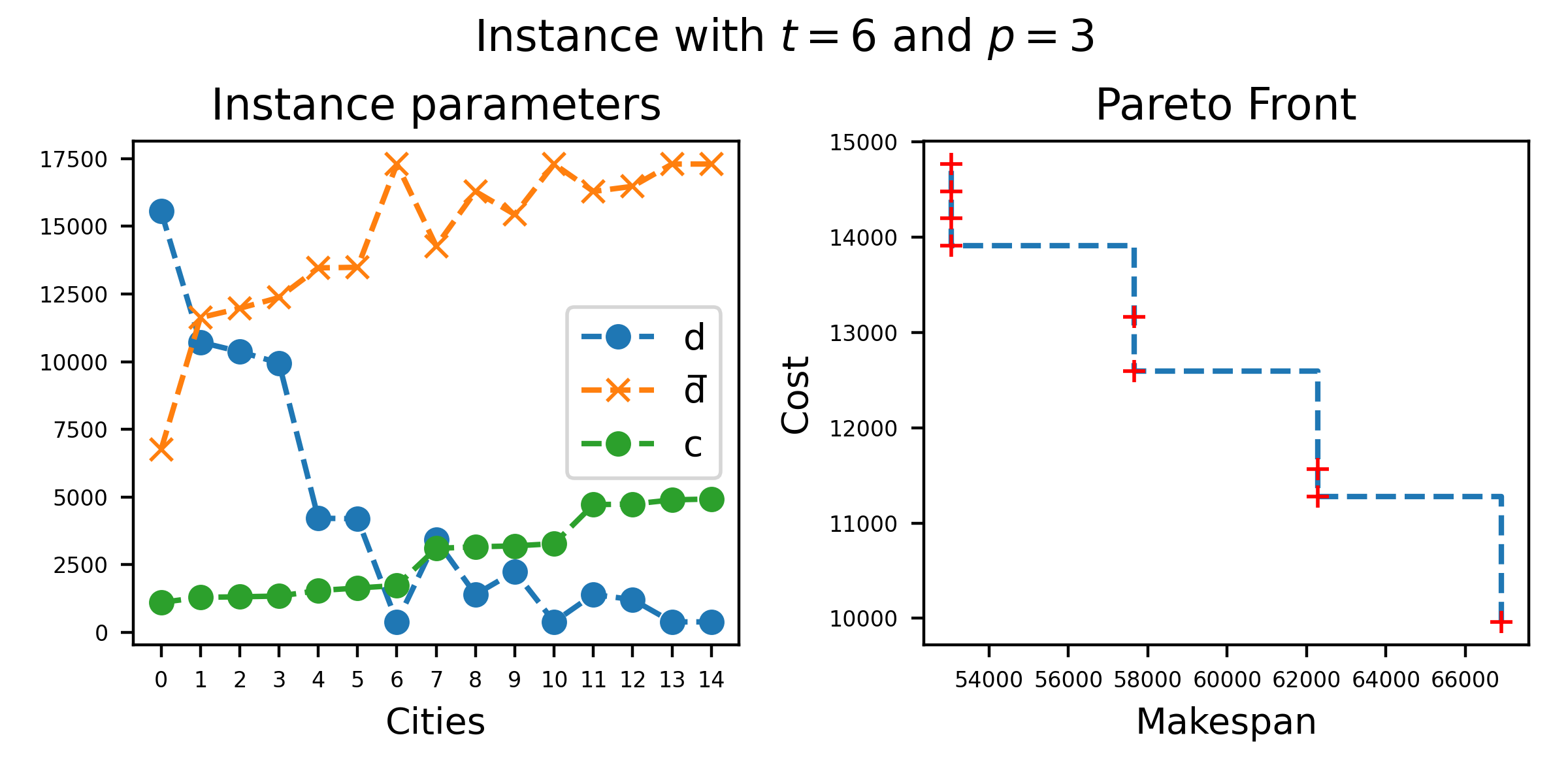}\hfill \\
      \includegraphics[width=0.65\textwidth]{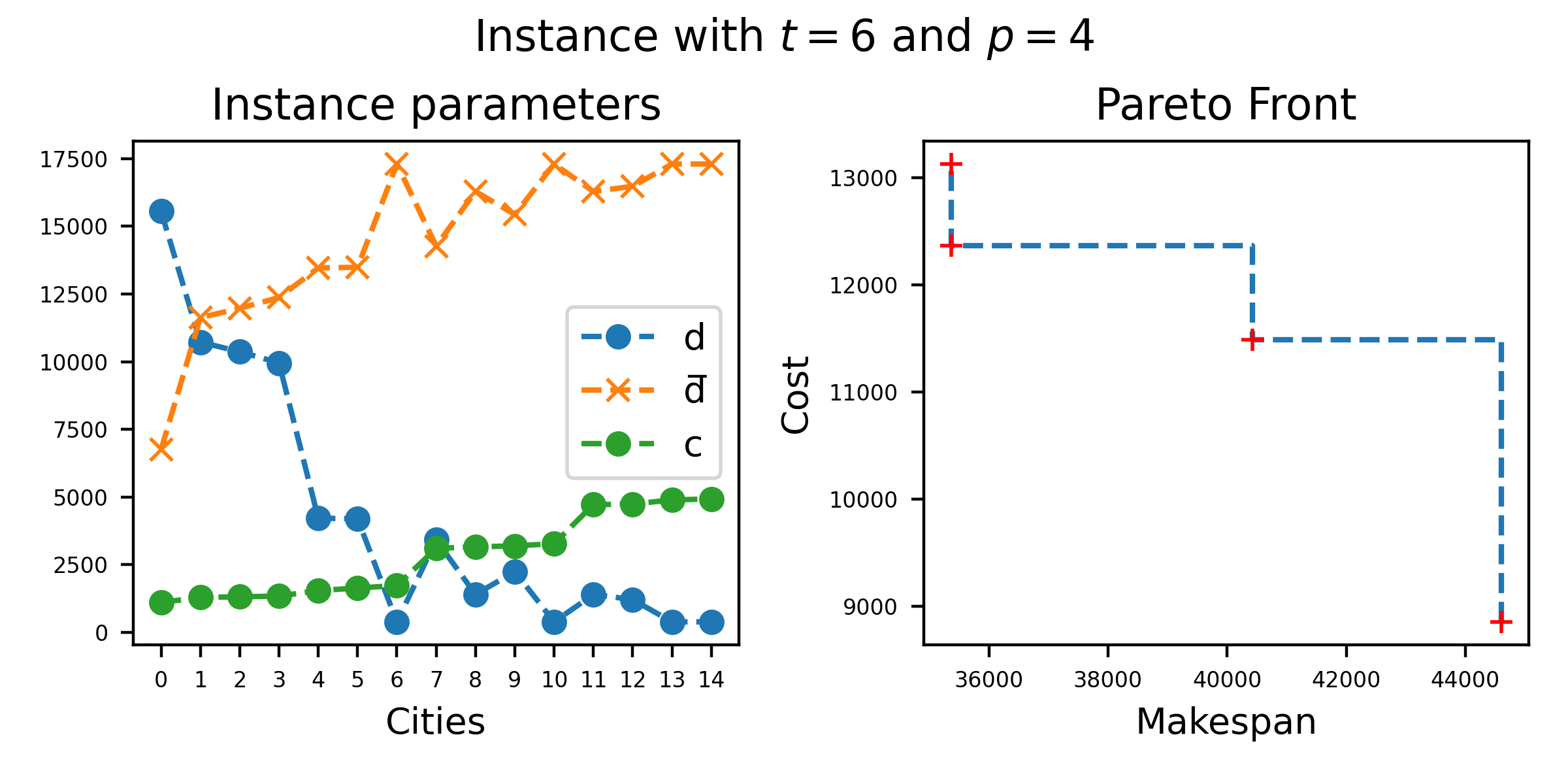}\hfill
      \includegraphics[width=0.65\textwidth]{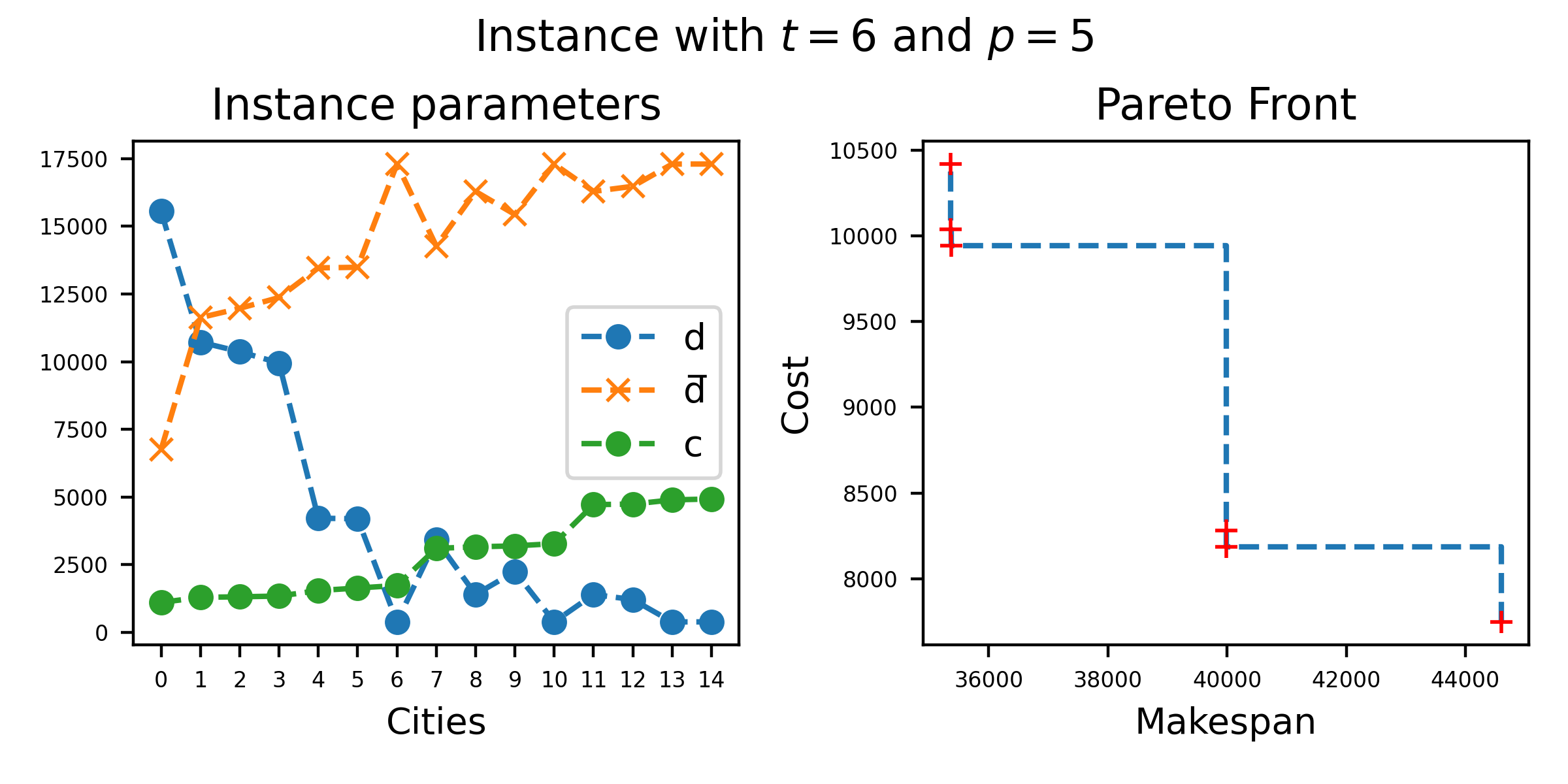}\hfill
 \caption{\label{openflight} Openflight instance with $n = 15$, $t=6$ and $p$ from 2 to 5.}
\end{figure}

The instance with 6 travelers and 2 planes has a Pareto front made of 29 distinct objective vectors. The two extreme points (Cost, Makespan) are $(111505, 11070)$ and $(88410, 16410)$. These two points are obtained by using exclusively one airport, either Dubai (DBX) or Seattle (SEA) and do not perform any $BA\bar B$ pattern:\\

\begin{verbatim}
C. 88410 Mk. 16410 
(5,5,5,5,5,5,5,5,)(5,5,){}
ATL -> SEA -> PEK -> SEA -> ATL -> SEA -> PEK -> SEA -> ATL -> SEA -> PEK
ATL -> SEA -> PEK -> SEA -> ATL -> SEA -> PEK -> SEA -> ATL -> SEA -> PEK
\end{verbatim}

~\\

\begin{verbatim}
C. 111505 Mk. 11070 
(0,0,0,0,0,0,0,0,)(0,0,){}
ATL -> DXB -> PEK -> DXB -> ATL -> DXB -> PEK -> DXB -> ATL -> DXB -> PEK
ATL -> DXB -> PEK -> DXB -> ATL -> DXB -> PEK -> DXB -> ATL -> DXB -> PEK
\end{verbatim}

~\\However, there are non-dominated path using other airports, notable San Francisco (SFO) and performing $BA\bar B$ patterns:\\

\begin{verbatim}
C. 97212 Mk. 14238 
(0,4,)(4,5,){0,4,4,}
ATL -> DXB -> ATL -> SEA -> PEK -> SFO -> PEK -> DXB -> PEK -> SFO -> PEK
ATL -> SFO -> ATL -> SFO -> ATL -> SFO -> PEK -> DXB -> PEK -> SFO -> PEK
\end{verbatim}

~\\Finally, for some Pareto optimal plan, planes do not always perform the same number of flights:\\

\begin{verbatim}
C. 101854 Mk. 13645 
(0,4,)(5,5,){0,0,5,}
ATL -> DXB -> ATL -> SEA -> PEK -> SFO -> PEK -> DXB -> PEK
ATL -> DXB -> ATL -> SEA -> PEK -> SEA -> PEK -> DXB -> PEK -> DXB -> PEK -> SEA -> PEK
\end{verbatim}

~\\Of course, in this example, the cost was set up arbitrarily and does not reflect any real cost but one can imagine that the cost is defined to represent some tax, or a sort of risk, i.e. linked with a certain infectious disease or an on-going conflict.

As expected, the front is composed of fewer points when the number of planes increases as reported in Figure \ref{openflight}. 

\section{Conclusion and Perspectives}
\label{sec:conclusion}

In this article, we extended our preliminary work \cite{quemy:hal-01109777} by relaxing the unrealistic assumptions. In particular, in this work, we only assume the triangular inequality to hold for the durations. First, we defined three types of \MULTIZENO~problem: the symmetric clique, the non-symmetric clique and the general version. 

The algorithm to identify and build Pareto optimal plans relies on a single proposition. Due to the increasing complexity of these problems, we first proved the proposition for the symmetric problem and then extended the proof for the non-symmetric version. For the general version, we showed that any instance can be reduced to a clique instance.

From an implementation point of view, we presented an optimized version of the \ZENOSOLVER~ which allows to tackle problem twice as big as the original version. We also implemented the non-symmetric version of the algorithm and demonstrated its performances and effect of pruning.

We demonstrated the diversity of Pareto-fronts which can be obtained by changing the instance parameters.
Finally, we provided a concrete application using real-life data. Using OpenFlight database, we used the \ZENOSOLVER~to find all the Pareto-optimal plans between the two largest airports in the world.

Beside the direct interest for the route and schedule multi-objective optimization for air transport, we believe that the work presented in this paper can be useful in many regards, and in particular for the benchmarking and comparison of algorithms, and for exploratory landscape analysis.
Moreover, existing multi-objective planing instances are, as far as we know, not offering the exact Pareto-front, which only allows the comparison between solvers but not to characterize how hard is an instance and how far from an optimal solution the solvers are.
On the contrary, \ZENOSOLVER~is capable to return the Pareto Front, with at least one plan for each point of the Pareto Front in the objective space.

Future work should focus on returning the entire Pareto set, i.e. all feasible plans for any Pareto optimal objective vector.
Another possible improvement would consist in characterizing the graphs for which Proposition IV holds,
because this would allow to know the \emph{general} \MULTIZENO~instances for which the reduction to a \emph{clique} \MULTIZENO~can be done in polynomial time.

\bibliographystyle{alpha}
\bibliography{bibliography,multizeno}

\end{document}